\documentclass[journal]{IEEEtran}
\usepackage{url}

\ifCLASSINFOpdf
\else
\fi
\usepackage{amsmath}

\usepackage{algorithmic}

\usepackage{subfigure}
\usepackage{graphicx}

\newcommand{\etal}{\textit{et al.}}
\usepackage{tabularx}

\usepackage{array}
\begin{document}
%
\title{Review Neural Networks about Image Transformation Based on IGC Learning Framework with Annotated Information}
%
%
%

\author{ Yuanjie Yan, Suorong Yang, Yan Wang, Jian Zhao, Furao Shen 
\thanks{Manuscript received April 15, 2021; revised xx, xx, xxxx; accepted xx, xx, xxxx. Date of publication
xx, xx, xxxx; date of current version April 15, 2021. This work is supported in part by the National Natural Science Foundation of China under Grant 61876076. The associate editor coordinating the review of this manuscript and approving it for publication was Prof. xx. (Corresponding author: Furao Shen and Jian Zhao.)
}
 \thanks{Y. Yan, H., S. Yang, Y. Wang and F. Shen are with the National Key Laboratory for Novel Software Technology, Department of Computer Science and Technology, Nanjing University, Nanjing 210046, China (e-mail: yanyj@smail.nju.edu.cn;  yangsr@smail.nju.edu.cn; wangy@smail.nju.edu.cn; frshen@nju.edu.cn ).}%
 \thanks{J. Zhao is with the School of Electronic Science and Engineering, Nanjing University, Nanjing 210046, China (e-mail: jianzhao@nju.edu.cn). He is also with the National Key Laboratory for Novel Software Technology, Nanjing University, China.}
}

%
%

\markboth{IEEE TRANSACTIONS ON PATTERN ANALYSIS AND MACHINE INTELLIGENCE}
{Shell \MakeLowercase{\textit{\etal~}}: Bare Demo of IEEEtran.cls for IEEE Journals}
%



\maketitle

\begin{abstract}

Image transformation, a class of vision and graphics problems whose goal is to learn the mapping between an input image and an output image, develops rapidly in the context of deep neural networks. In Computer Vision (CV), many problems can be regarded as the image transformation task, e.g.,  semantic segmentation and style transfer. 
These works have different topics and motivations, making the image transformation task flourishing. Some surveys only review the research on style transfer or image-to-image translation, all of which are just a branch of image transformation. 
However, none of the surveys summarize those works together in a unified framework to our best knowledge. 
This paper proposes a novel learning framework including  Independent learning,  Guided learning, and  Cooperative learning, called the IGC learning framework. The image transformation we discuss mainly involves the general image-to-image translation and style transfer about deep neural networks.
From the perspective of this framework, we review those subtasks and give a unified interpretation of various scenarios. 
We categorize related subtasks about the image transformation according to similar development trends. Furthermore, experiments have been performed to verify the effectiveness of IGC learning. Finally, new research directions and open problems are discussed for future research. 

\end{abstract}


%
\IEEEpeerreviewmaketitle

\section{Introduction}
%
%
%
%

\IEEEPARstart{D}{igita}l image transformation, which has a long history in the field of computer graphics and computer vision, is to transform the original image into another form according to the needs. 
Many mathematical methods, such as image rendering [3] and image spatial filters [4], are traditional image transformation algorithms. 
However, these methods can not handle more complex and creative tasks. 
Image transformation using deep neural networks, also known as image-to-image translation, can complete more sophisticated and artistic tasks, such as semantic segmentation \cite{Shelhamer2017Fully} and style transfer \cite{Gatys2016Image}. 
Semantic segmentation can be regarded as an image transformation task from the original image to a semantic segmentation mask. 
The development of semantic segmentation has an important influence on image transformation with neural networks. Style transfer is the conversion of the content image to an image with another style, which is a problem of art and re-creation. 
These two typical image transformation scenes are traditionally studied independently in context of neural networks. 

\begin{figure}[t]
\begin{center}
\includegraphics[scale=0.4]{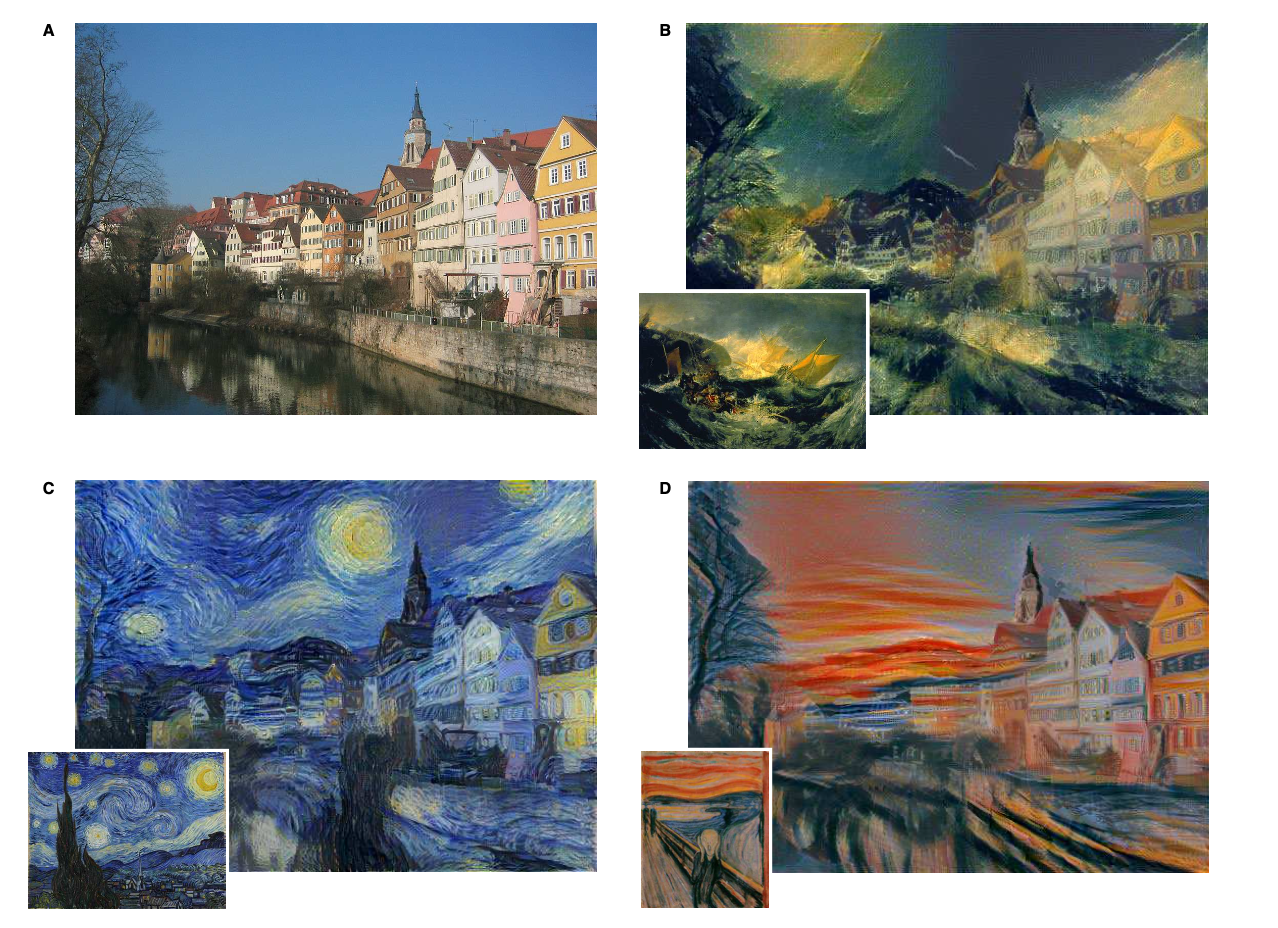}
\caption{A example of image style transfer from \cite{7780634}. A) The original photograph depicting the Neckarfront in T{\"u}bingen, Germany (Photo: Andreas Praefcke). The painting that provided the style for the respective generated image is shown in the bottom left corner of each panel. B) \textit{The Shipwreck of the Minotaur} by J.M.W. Turner, 1805. C) \textit{The Starry Night} by Vincent van Gogh, 1889. D) \textit{Der Schrei} by Edvard Munch, 1893.}
\label{style_image}
\end{center}
\end{figure}

Deep learning has made breakthroughs in many sophisticated scenes, such as Computer Vision (CV) \cite{He_2016,ren2015faster,Chou2020360IndoorTL}, and Neural Language Process (NLP) \cite{devlin2018bert_v2,brown2020language}. 
With the development of deep learning, it is common for a neural network to have millions or billions of parameters, which makes it more and more difficult to train. 
Building large labelled training data sets is a good method to overcome the notorious over-fitting problem in machine learning. 
In CV,   ImageNet \cite{5206848} and MS COCO\cite{Everingham2009ThePV} are constructed and published for image classification \cite{He_2016}, semantic segmentation \cite{ren2015faster}, instance segmentation \cite{He_2017}. In NLP, General Language Understanding Evaluation (GLUE) \cite{Wang2018GLUE} also summarizes nine datasets for different subtasks about neural language.
	 However, it takes a lot of time and effort to label data manually. 
In addition, it is impossible to label the dataset manually for some CV tasks, such as style transfer which is one of the themes of this article as shown in Fig.~\ref{style_image}. 
There is no clear label for style transfer, but it is necessary to generate an image which combines content and style images.
Therefore, learning neural network model from unlabelled data or a small number of labelled samples is the current research trend.
On the other hand,  some self-supervised methods of neural networks can utilize the unlabelled dataset in special tasks, e.g., Generative Adversarial Network (GAN) \cite{radford2015unsupervised} and adversarial examples generation \cite{8294186}.
GAN maps the random vector to a specific image via the discriminator network to judge the true and false of the generator output. In adversarial examples generation, the adversarial image is iteratively derived from the target network model about the original image. 
In the above two methods, other models are used to label the unlabelled data, so as to realize the target network training.
Besides, meta-learning \cite{hospedales2020metalearning} and few-shot learning \cite{bendre2020learning} are also proposed to deal with the problem of fewer labelled samples.

The image-to-image translation is a broad issue in the literature of CV and graphics where many subtasks have been researched individually.
We will not discuss all the work of image-to-image translation but will 
reveal the relationships between different tasks with a unified perspective in this article.
Therefore, we mainly discuss general image-to-image translation tasks and specifically the subtask about image style. 
The development of image style transfer is related to the development of CV. Gatys \etal~\cite{7780634} first demonstrates the power of Convolution Neural Networks (CNNs) to render a natural image with different painting styles in Fig. \ref{style_image}. 
They apply the pre-trained CNN model to analyse the image content and image style as the annotated label. 
The annotation information from the model is different from that annotated  manually.
The work of Gatys \etal~not only opens up a new field called style transfer (NST) but also demonstrates a way where a neural network can learn from another pretrained model without manual labels. Since then, there have been lots of follow-up studies to improve the NST methods from various aspects in academia \cite{Jing_2020} and many successful industrial applications in industry \cite{deepforge}.
In addition, the concept of neural image style transfer has been extended to the field of NLP.
Style transfer in NLP represents variations of a text without modifying its meaning \cite{fu2017style}. 

In this paper, we aim to provide a new learning framework based on annotated information.  We mainly discuss the image transformation methods using neural networks, referred to as image transformation for short.
Such a learning framework is called IGC learning, which comprises three learning paradigms, namely independent learning, guided learning, and cooperative learning.
Note that those learning paradigms are related to other learning paradigms presented in other works \cite{8237558,Yu_2020}, but we provide a unified definition based on the annotated information of labels.
In addition, we review and analyse some representative works about image transformation from the perspective of the IGC learning framework. 
Then, we use the same method to organize image transformation and style transfer.
We show that there is a close relationship between the development of image translation and style transfer. 
In order to verify the effectiveness of each learning paradigm, we perform experiments and compare related methods on various datasets. 
What's more, we also introduce related applications of image transformation.
Finally, we discuss some problems of image transformation, and extend the IGC learning framework to other areas. 
The main contributions of this paper can be summarized as follows:
 \begin{itemize}
 	\item We propose a new learning framework called IGC learning in term of the annotated information. 
 	\item From the perspective of the IGC learning framework, we analyse representative models and categorize more works in research of image transformation about neural networks.
 	\item We first demonstrate that the development of image-to-image translation is highly related to the development of style transfer.
 	\item We discuss open problems of the image transformation and extend the IGC learning framework for future research.
  \end{itemize}

The rest of the paper is organized as follows.
First, we introduce the background of image transformation and style transfer in Section \ref{background}.
Then, we propose the novel IGC learning framework in detail in Section \ref{igc learning}. Section \ref{image style transfer} reviews some representative image transformation models from the perspective of the IGC learning framework. In Section \ref{more_methods}, we categorize more works in the research of image transformation and survey their motivations and contributions. In Section \ref{experiment}, we perform relevant experiments and discuss the learning methods in IGC. The applications of image transformation are introduced in Section \ref{application}. 
Moreover, we discuss open problems in image transformation and extend our proposed IGC learning framework in Section \ref{furture}.
We delineate several promising directions for future research.
Finally, Section \ref{conclusion} summarizes our themes and contributions in this article.

\section{Background of Image Transformation}
\label{background}

The goal of image transformation is to learn a mapping from an input image to an output image.
In the image semantic segmentation task, the each pixel of the original image is classified according to whether it belongs to the same object.
Aligned image pairs are required to train the network. With the help of sufficient aligned data, the methods about semantic segmentation based on neural networks surpass other traditional algorithms, e.g., Conditional Random Field (CRF)~\cite{7780720} to achieve state of the art. Semantic segmentation can be seen as a special task of image transformation on aligned data.

However, the cost of a large-scale aligned image dataset is tremendous. Inspired by GAN, some works are proposed to solve the lack of aligned images. To a certain extent, 
researchers construct a new problem about the image transformation on two unaligned datasets. 
As the flexibility of GAN, there are many studies to follow up, e.g., DualGAN \cite{Yi2017DualGAN} and UNIT \cite{liu2017unsupervised}. 
These methods as variants of GAN can solve the general image transformation problems. 
Besides, some interesting research has been derived by combining the image transformation and objection detection. Researches introduce conditional GAN (cGAN) \cite{mirza2014conditional} models to control or modify partial areas in image. Partial image transformation (e.g., face swapping~\cite{TOLOSANA2020131} and lip swapping~\cite{ Agarwal_2019_CVPR_Workshops}) have attracted great attention in academia and also produced popular applications in industry. 
Those methods also work on unaligned datasets.

\begin{figure*}[t]
\begin{center}
\includegraphics[scale=0.42]{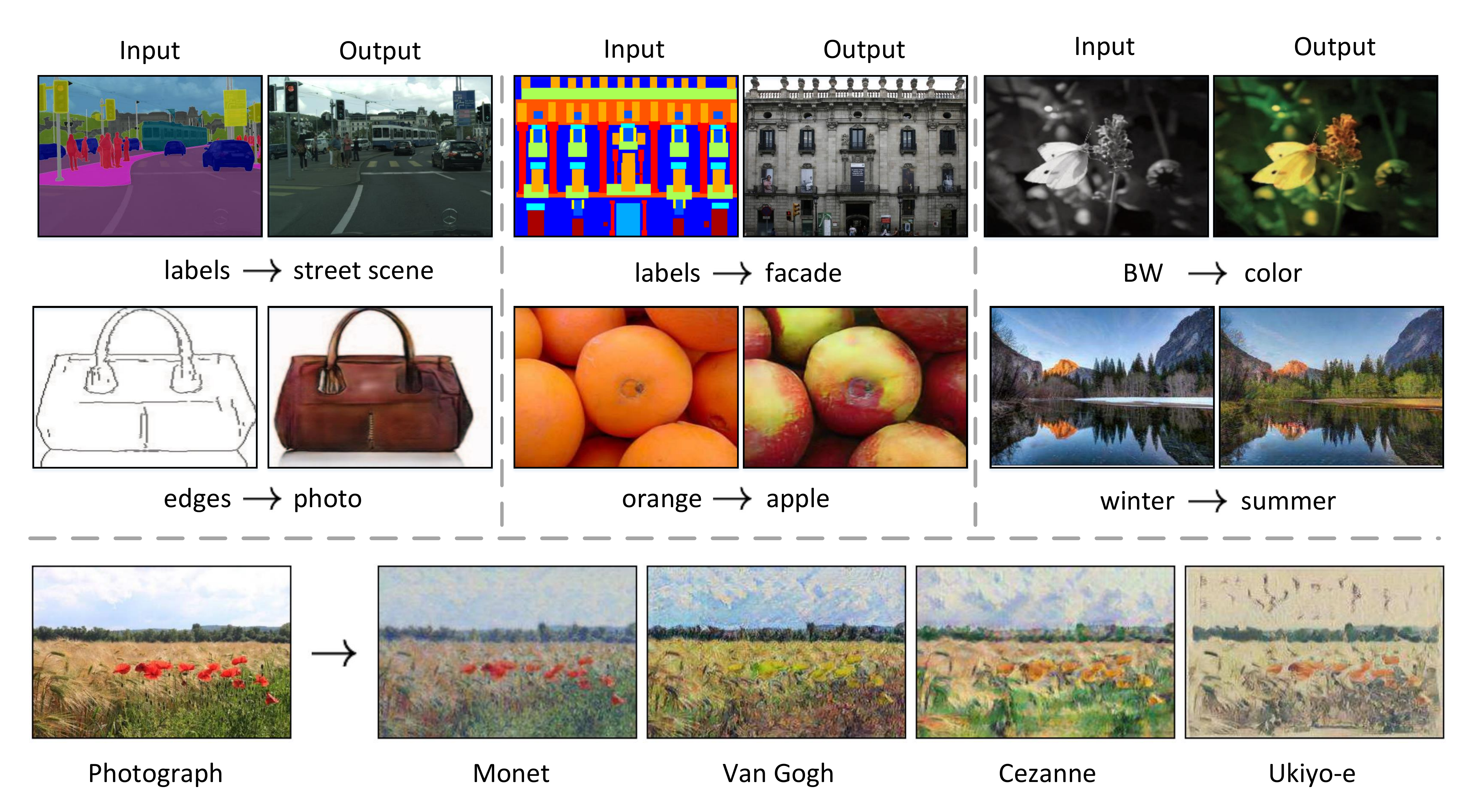}
\caption{The general image transformation. The sample cases belong to the aligned dataset in the first line. The second line contains the unaligned sample of image transformation. The case in the third line shows the multiple transformations from one original image.}
\label{general image transformation}
\end{center}
\end{figure*}

Furthermore, style transfer, one sub-task of the image transformation, accomplishes an artistic image-to-image translation to repaint a content image with a kind of style from another image. This method exploits the pre-trained neural network model to extract the image feature and style as supervised information. Consequently, there is no need  for tagging data or even datasets. Only a content image and a style image are needed. It is very different from the aforementioned methods of image transformation by GAN.

\begin{figure*}[t]
\begin{center}
\includegraphics[scale=0.6]{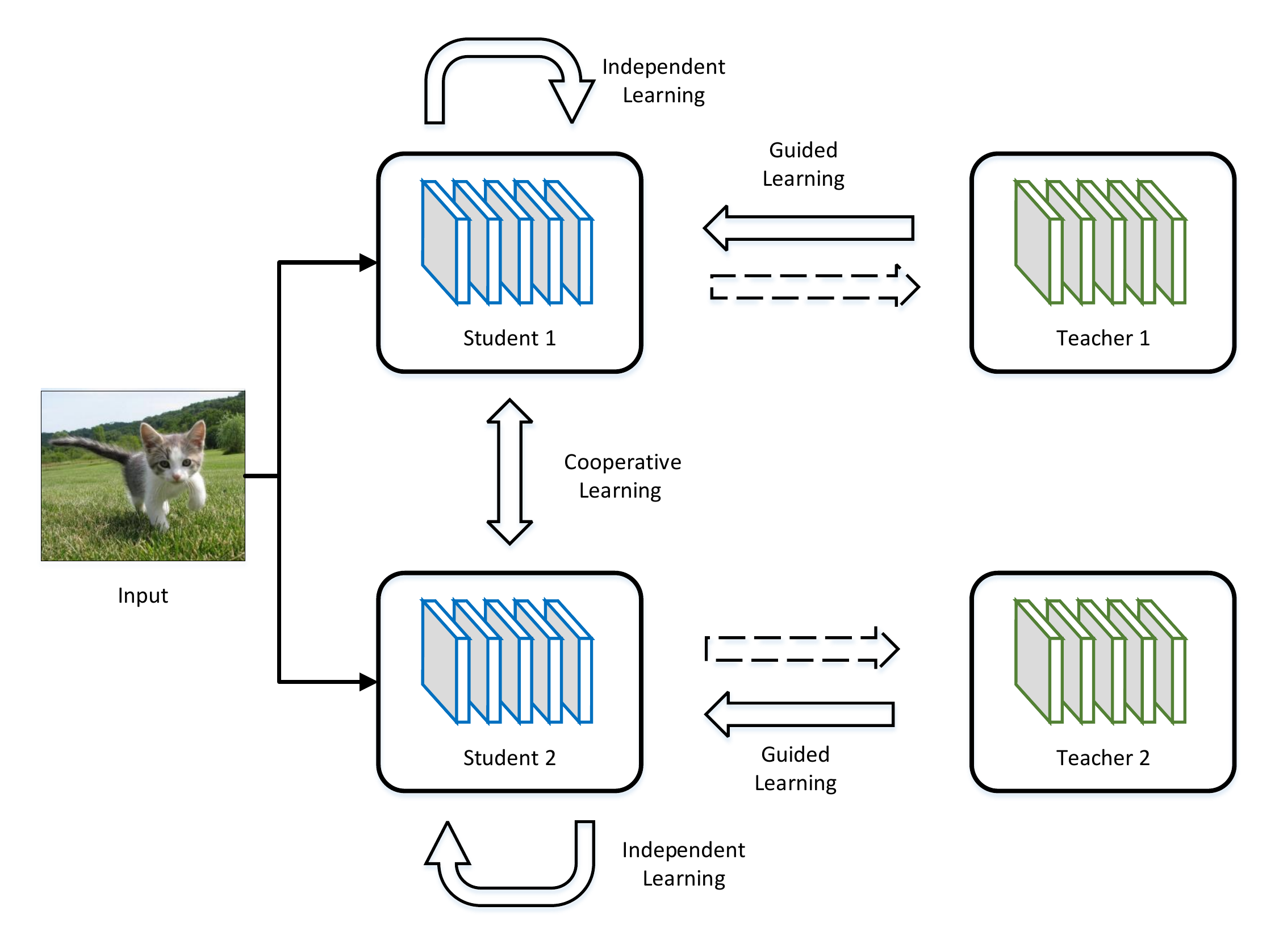}
\caption{The IGC learning framwork. The figure only shows independent learning, guided learning between a single student model and a teacher model, and cooperative learning between two student models.}
\label{igc}
\end{center}
\end{figure*}

\begin{table*}[t]
\caption{Datasets about image transformation tasks}
\begin{center}
\begin{tabular}{c|c|c|c|c}
\hline
Tasks & Dataset  A & Dataset B & Pattern & Description \\
\hline 
Semantic segmentation 		& Semantic labels & Photo & Aligned & from Cityscapes dataset \cite{2016The}\\
Semantic segmentation:		& Architectural labels & Photo & Aligned & from CMP Facades dataset \cite{2013Spatial}\\
Colourization		& Black and white picture & Colour photos & Aligned & from dataset \cite{5206848}\\
Photo generation & Edges & Photo & Aligned & from \cite{Zhu_2016} \\
\hline 
Object transfiguration &  the class images of object A& the class images of object B & Unaligned & the object class from ImageNet \cite{5206848} \\
Season transfer  & Winter photos of Yosemite & Summer photos of Yosemite & Unaligned & collected in \cite{Zhu_2017}. \\
Style transfer & the content dataset & the style dataset & Unaligned & from \cite{Gatys2016Image} \\
\hline
Style transfer & the content image & the style image & Two images & from \cite{Gatys2016Image}\\
Style transfer & the content dataset & the style image & Dataset \& image & from \cite{Zhu_2017} \\
Style transfer & the content dataset & the style image dataset & Unaligned & from \cite{luan2017deep} \\
\hline
\label{dataset}
\end{tabular}
\end{center}
\end{table*}
 
To summarize the datasets on image transformation, 
we classify the three dataset types:
aligned  between two datasets, unaligned  between two datasets and only two image examples without datasets. 
Different types of datasets are built for different granularities of image transformation tasks.
The aligned dataset is for image transformation tasks with pixel-level granularity. The unaligned dataset is aimed at target-level granular image transformation tasks. Only two image examples without datasets are often referred to as style transfer, which relates to perception by a human from the image-level granularity.
In Table \ref{dataset}, we summarize some datasets about image transformation tasks. 

Currently, most works focus on the unaligned datasets. There are two main advantages. First, compared with aligned datasets, unaligned datasets are easier to create. 
Second, in the context of unaligned datasets, the ability of image transformation methods is flexible and its application is wider. 
We define these works as the general image transformation. 
For style transfer, due to the discrepancy among the above methods, it has also been studied individually. 
Consequently, we mainly focus on general image transformation and style transfer in this paper. 

\subsection{General Image Transformation}
\label{general}
General image transformation contains various image-to-image translations on aligned and unaligned datasets as shown in Fig. \ref{general image transformation}. 
General image transformation mostly focuses on the objects in the image.
Therefore, it is crucial to identify and locate objects.  Image classification and object detection as the basic problems in CV have been researched widely.
Some popular neural networks have been proposed like ResNet \cite{He_2016} and VggNet \cite{Zhang_2016}.
The architecture of those network has been adopted and applied in a wide range of image problems.
Follow those works, most of image transformation methods use similar networks to build transformation models. At the beginning of the research on image-to-image translation, researchers have directly applied the semantic segmentation models to aligned datasets. Then, the general image transformation on the unaligned dataset has been proposed, which has attracted widespread attentions. 
In this paper, we discuss the general image transformation on aligned and unaligned datasets. The solutions on the two types of datasets have similar network models but there are discrepancies in learning methods.

\subsection{Style Transfer}
The image style comes from subjective feelings by humans and has no clear definition. The image style is related to color, texture, lighting, etc.  
The traditional method mainly adopts spatial filter, which achieves various styles by defining transformations on RGB channels, e.g., comic strip filter \cite{park2011anatomy} and nostalgia filter \cite{bartholeyns2014the}. 
Since most image styles can not be expressed explicitly, we describe the style using a style image.

The work of Gatys \etal~utilizes a pretrained convolution neural network to extract the style of images and achieve fusion images like Fig. \ref{style_image}.
Since then, many works about style transfer have been proposed from different perspectives. Survey \cite{Jing_2020} categorizes those works according to the image-optimization-based online neural method and model-optimization-based offline neural method. The image-optimization-based online neural method means that it generates the target image iteratively with the original image and the style image.  
The model-optimization-based offline neural method means that it uses the transform neural network, which has been trained for a certain style before, to generate the target image directly from the original image. However, it did not reveal the relationship between the two methods.  In this survey, we try to adopt a unified architecture to demonstrate their relationship.

\section{IGC Learning Framework} 
\label{igc learning}

With the development of hardware, we can solve more complex problems by designing larger neural network models which contains millions or billions parameters. To avoid the overfitting problem, a large training dataset is needed  in the corresponding large model. However, a large amount of data from the internet is unlabelled. Labelling data manually is tedious in time and cost, even impossible in some tasks like the image style transfer. Fortunately, some neural network models can use other models to annotate data. Inspired by their works, we summary and propose a new learning framework called IGC learning, which comprises three learning paradigms, namely Independent learning, Guided learning, and Cooperative learning. 
Single or multiple models can be included under the IGC learning framework.
According to whether the model can be learned, it is divided into student model and teacher model, 
which is consistent with the relevant description in knowledge distilling \cite{Hinton2015Distilling}.

We define the following mathematical symbols about the model in the IGC learning framework.
\begin{itemize}
\item Student model: we denote $S^i_w$ as the $i$-th student model which is needed to learn parameters $w$. 
\item Teacher model: we denote $T^i$ as the $i$-th teacher model which does not need to learn the parameters.
\end{itemize}

For specific problems, we can construct any learning model and decompose it into several student models and teacher models in the perspective of the IGC learning framework.
Then, 
depending on how the data is labelled, we define three learning paradigms about the IGC learning.

\begin{itemize}
\item Independent learning is the learning process of the student model based on data.
\item Cooperative learning is the learning process among some student models. 
\item Guided learning is the learning process between the student models and teacher models. 
\end{itemize}

Fig.\ref{igc} shows the IGC learning framework with two student models and two teacher models.
We believe that the IGC learning framework is more suitable for neural network models than the mainstream supervised and unsupervised learning frameworks.
First, multiple models are often involved in a complex image task and the IGC learning framework is capable of analysing their relationships. Second, the IGC learning framework guides us to build more suitable models by introducing new models and learning methods.
Note that IGC framework is mainly proposed for neural network models, but it is not limited to it. We can also apply this framework to the analysis of the combination of traditional models and neural network models \cite{9062501}.
Furthermore, since the three learning paradigms are independent from each other, the student model can perform three learning methods simultaneously. 
Each learning method will improve the performance of the student model to a certain extent. 
We discuss more details of the learning paradigm about the IGC and give some examples of related models.


 

\subsection{Independent Learning}
\begin{figure}[t]
\begin{center}
\includegraphics[scale=0.33]{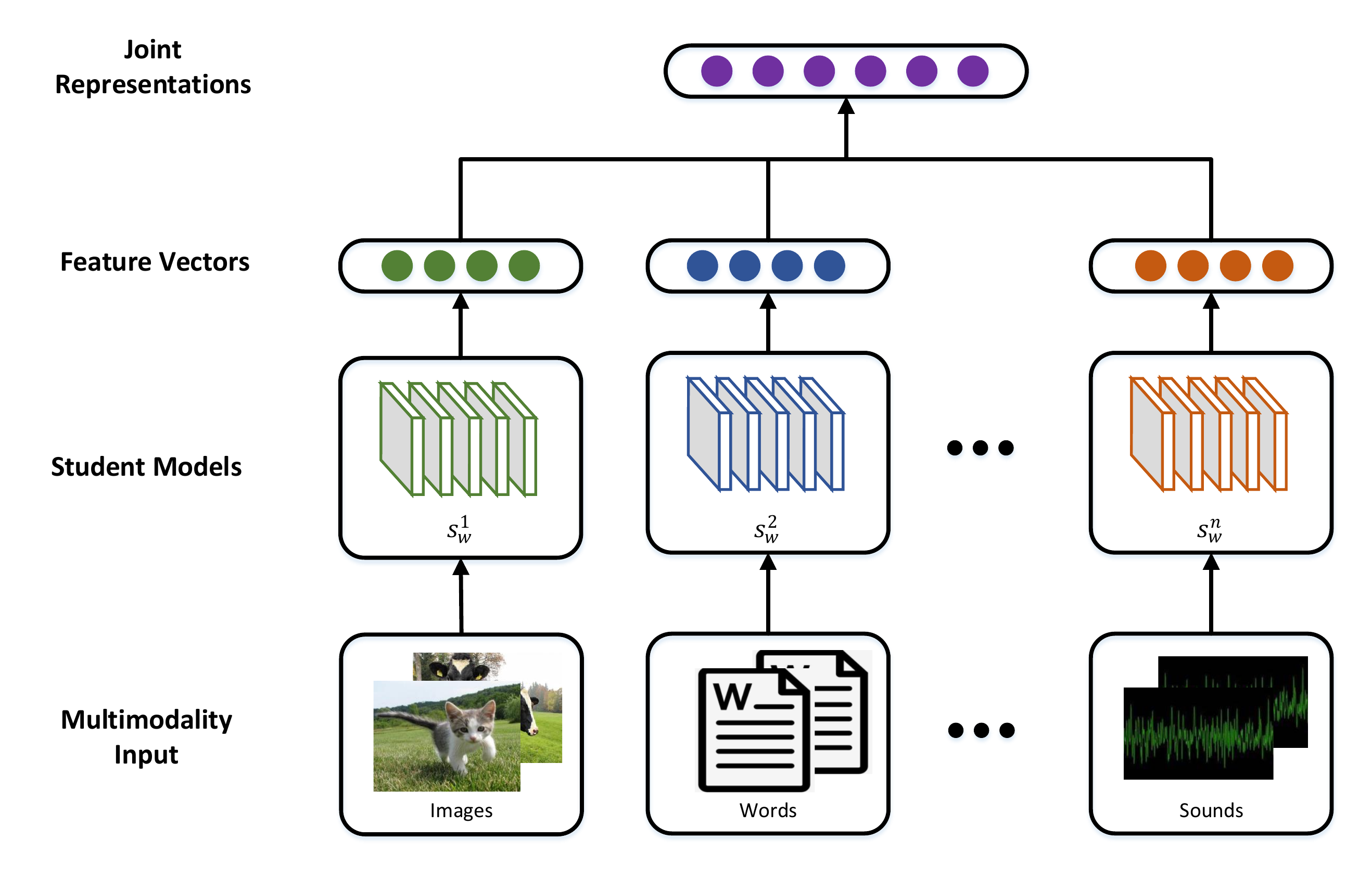}
\caption{The framework of joint representations in multimodal learning}
\label{multimodel}
\end{center}
\end{figure}

Independent learning is the mainstream way in deep neural networks. When there is only one model, we denote it as $S_w$. For most  neural networks, the student model $S_w$ is trained on the labelled dataset. 
Multimodal learning \cite{Ngiam_multimodaldeep} involving a variety of networks is to improve the performance of models via the cooperation on some related models. 
Joint representations \cite{mao2017deepart} is a typical multi-model method. 
As shown in Fig. \ref{multimodel}, the model can divide some student models that denote as $S^1_w, S^2_w,\cdots, S^n_w$, where $n$ is the number of networks. 
Multimodal learning just using the manually labelled data belongs to independent learning.
Self-supervised learning methods that are different from supervised learning,
e.g., Autoencoder\cite{Chen2016Variational} and BERT\cite{devlin2018bert}, also belong to independent learning in the context of the dataset. The annotations of data are not from other models but from the data itself.
Therefore, independent learning is not just a subset of supervised learning. 

\subsection{Cooperative Learning} 
\label{cooperative learning}
\begin{figure}[t]
\begin{center}
\includegraphics[scale=0.4]{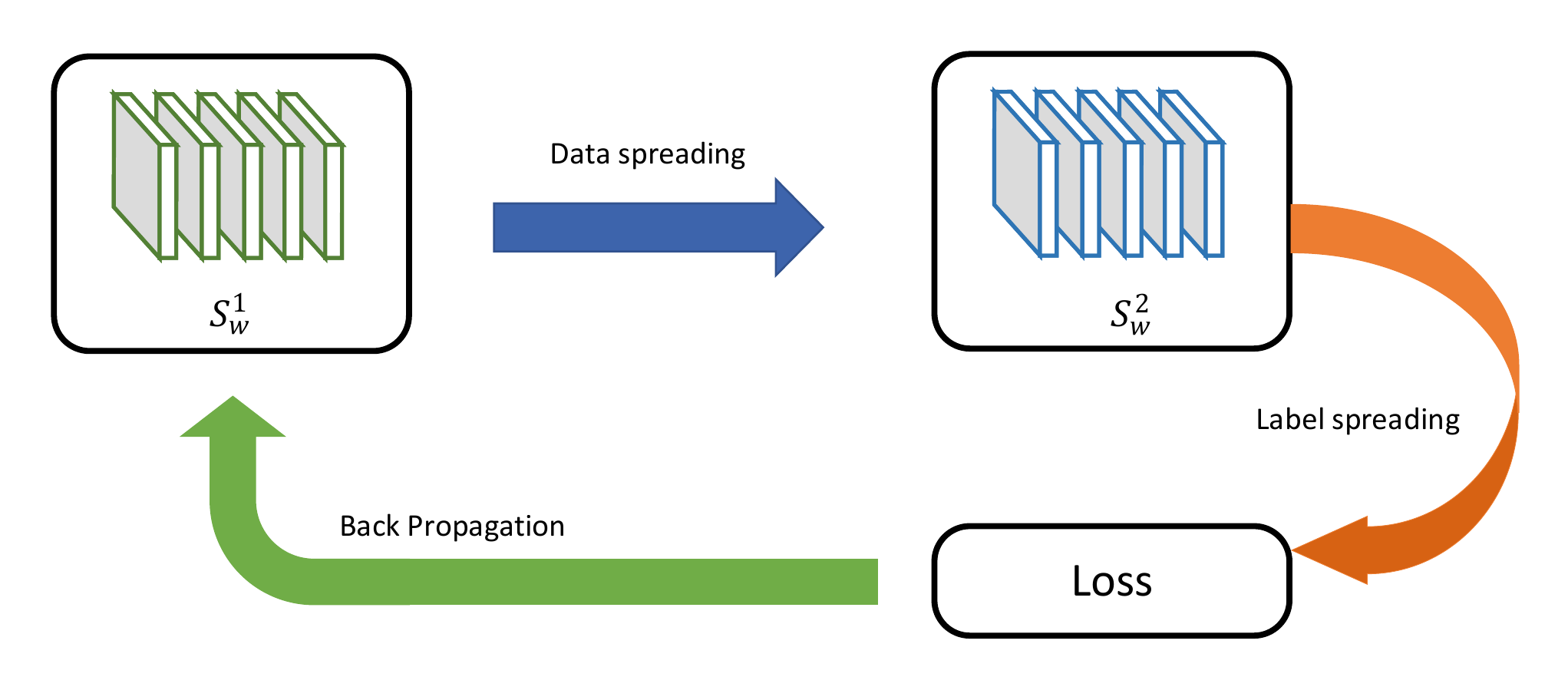}
\end{center}
\caption{ Data spreading and label spreading between $S^1_w$ and $S^2_w$.}
\label{cooperative}
\end{figure}

Cooperative learning is the learning process among some student models. 
The purpose of cooperative learning is to share knowledge or confront each other among student models to improve their ability.
GAN is most relevant to the cooperation learning we proposed. The typical GAN model has a generator model and a discriminator model, which can be seen as student model $S^1_w$ and $S^2_w$, respectively. In the image generation task, the $S^1_w$ model maps random vectors to real images, while the $S^2_w$ model classifies images as real images or generated images. 
The original adversarial loss and its variants \cite{2017Wasserstein} are proposed to train the models $S^1_w$ and $S^2_w$. 
The Conditional Generative Adversarial Networks \cite{mirza2014conditional} replace random vectors with the the limited input to control the generated output as the student model $S^1_w$.  Most image transformations adopt the architecture of conditional GAN to accomplish the reconstruction from the source image to the target image.
Besides GAN, Dual Learning \cite{10.5555/3157096.3157188} where two models learn their parameters through duality tasks is also a typical cooperation learning method. 
The way of cooperative learning is to accomplish the spreading of data and loss among models.
As shown in Fig. \ref{cooperative}, for $S^1_w$, it communicates with $S^2_w$ to learn its parameters through data spreading and label spreading as follows,
\begin{itemize}
\item Data spreading means that the outputs of $S^1_w$ are the inputs of the $S^2_w$. 
\item Label spreading means that the  outputs of $S^2_w$ are the labels as the outputs of $S^1_w$. 
\end{itemize}

The learning of  $S^2_w$ is not related to any other models.
We call $S^1_w$ as the main student and $S^2_w$ as the associate student. 
For GAN, the generator is the main student and the discriminator is the associate student. 
In the next Section, we analyse some models that make full use of these two cooperative methods to strengthen the learning among students. 

\subsection{Guided Learning}
Guided learning is rather related to independent learning. Teacher model $T$ can come from the student model $S$ after independent learning, but it is not limited to this. The paper \cite {9062501} uses the traditional clustering model as the teacher model. 
Moreover, the teacher model $T$ is also called the pre-trained model in the fine-tuning method \cite{THORVALDSEN2020110352}. However, unlike the fine-tuning method, we try to learn knowledge from the pre-trained model instead of adjusting it. Knowledge distillation \cite{chen2020neural}, an effective way of neural network compression, is a typical guided learning. The task of knowledge distillation is to train a small network from the original big network but maintain the accuracy of the original model. 
Another important research of guided learning is feature visualization~\cite{Zeiler_2014}. 
The typical feature visualization method is to iteratively modify the random noise using the image features extracted from the teacher model $T$, which can be seen as the source of the image style transfer. 
Guided learning is also similar to the cooperative learning. Student models learn from teacher models through data spreading and label spreading. Teacher model can be seen as an excellent student model without learning. 

\section{Review Representative image transformation Methods based on the IGC Learning Framework}
\label{image style transfer}

Our proposed IGC learning framework is based on the annotation from datasets or models. This framework does not focus on the details of models, but on the learning methods among different models. 
From the perspective of the IGC learning framework, we analyse some representative works on image transformation. 

\subsection{Independent Learning of image transformation}
\label{semantic method}
\begin{figure}[t]
\begin{center}
\includegraphics[scale=0.48]{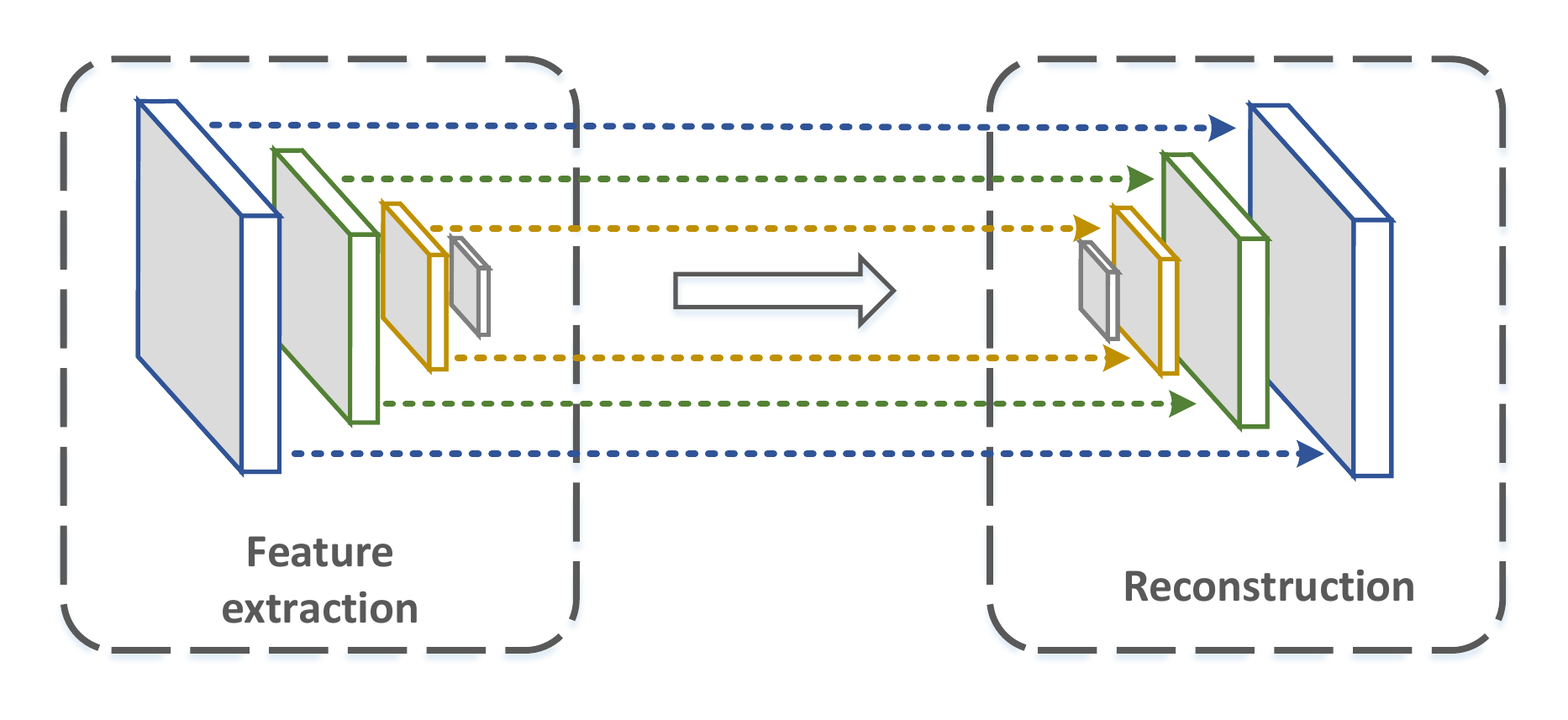}
\caption{A general model of semantic segmentation. The feature extraction can be seen as an encoder to extract various objects in the image. The reconstruction is to realize the mapping from object feature to mask. U-Net improves the accuracy of position for each pixel through skip layer connection (the dotted line in the figure).}
\label{semantic}
\end{center}
\end{figure}

Independent learning works on the aligned datasets. Much work of the image transformation on the aligned dataset is related to semantic segmentation. 
Various aligned datasets have been established to solve practical problems in Table \ref{dataset}. A large number of models have been proposed, e.g., Fully Convolutional Networks (FCN) \cite{Shelhamer2017FullyCN} and U-Net \cite{u_net}. They have a similar structure shown in Fig. \ref{semantic}. 
FCN adopts a basic autoencoder to map images to semantic masks through the downsampling to the upsampling.
U-Net also provides an easy but efficient method to locate the pixels of the mask by adding some skip connections from feature extraction to reconstruction. 
Based on the IGC learning framework, the student model denotes as $S_w$ to learn the parameters $w$ from the labelled dataset.   
$S_w$ is trained using a per-pixel loss between the output and ground-truth image. 
A per-pixel loss is defined as follows,
\begin{equation}
\mathcal{L}_{pixel}(S_{w}; x, y) =  (S_w(x) -  y)^2
\label{L2dis}
\end{equation}
where $x$ means a image and $y$ is the semantic mask of $x$.  

Although we do not care much about the details of the model under the IGC learning framework, a well-structured model is an important research direction related to semantic segmentation.
Various efficient models about independent learning are proposed in the literature of semantic segmentation in survey \cite{LATEEF2019321}. We only present two methods for semantic segmentation: FCN and U-Net, which are also adopted by other models that will be introduced later.

\subsubsection{FCN}
FCN is the first model to achieve an end-to-end and pixel-to-pixel neural network model in semantic segmentation. 
FCN is made up of fully convolutional layers that can take the input with arbitrary size and produce correspondingly-sized output. Using the upsampling convolution, FCN achieves the feature extraction and reconstructs the mask with the same size as the original image.


\subsubsection{U-Net}
U-Net is a kind of convolution networks for biomedical image segmentation.  It is widely adopted for simplicity but efficiency in small datasets. Compared with FCN, the biggest difference is that U-Net uses skip connections between feature extraction and mask reconstruction, which greatly enhances the accuracy of pixels after reconstructing the mask.


\subsection{Cooperation Learning of image transformation}
\label{cooperation learning}
Consider the opposite problem of semantic segmentation, converting the semantic mask into a realistic image. This problem can be solved by previous independent learning methods. However, the researchers found that the generated images are vague due to a lack of details. Inspired by GAN which can generate realistic and vivid images, more and more image-to-image translation tasks adopted the conditional GAN. As mentioned in Section \ref{cooperative learning}, conditional GAN is considered a kind of cooperative learning under the framework of IGC learning. At the beginning of the image transformation, the works of cooperative learning are aimed to the aligned dataset. Compared with independent learning, cooperative learning has significantly improved the details of the generated images, e.g., colourization and edge to photo in Table \ref{dataset}. An aligned dataset restricts the applications of the image transformation. Some researchers pay attention to accomplishing the object transformation of  images on unaligned datasets.
The survey \cite{8195348} summarizes the methods for two types of datasets in the context of GAN.

In this section, we analyse typical image transformation methods about cooperation learning in the IGC learning framework. 
We review the PixeltoPixel \cite{Isola_2017} on aligned dataset and CycleGAN \cite{liu2017unsupervised} on unaligned dataset.

\begin{figure}[h]
\begin{center}
\includegraphics[scale=0.25]{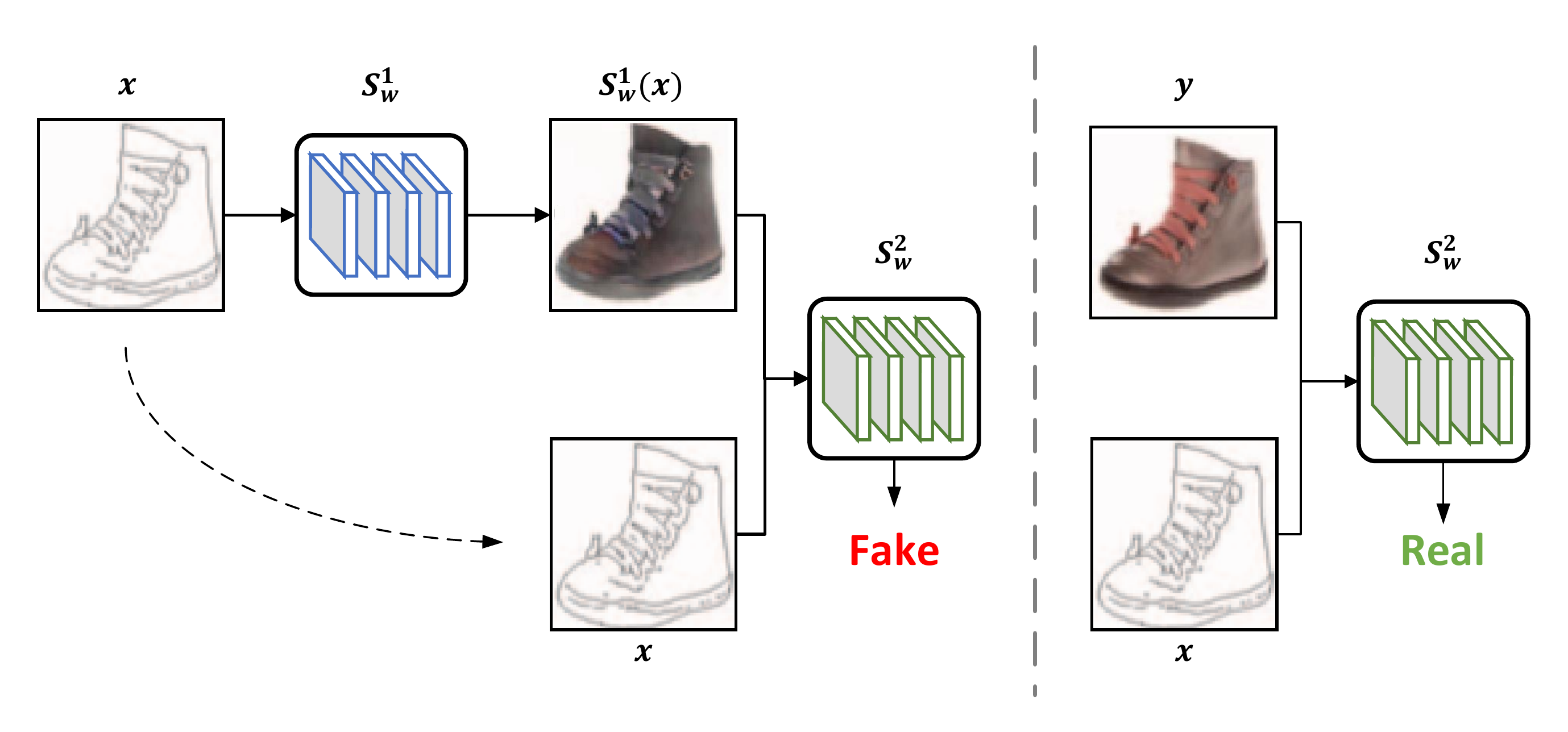}
\caption{Review the Pixeltopixel model based on IGC learning framework}
\label{pixeltopixel}
\end{center}
\end{figure}

\subsubsection{PixeltoPixel}
PixeltoPixel is a model to solve general image transformation problems in Fig. \ref{general image transformation}.  PixeltoPixel, a typical conditional GAN, has a generator as image transformation network and an discriminator as a classifier. As shown in Table \ref{dataset}, PixeltoPixel solves some transformation tasks on aligned dataset.

Based on the IGC learning framework, PixeltoPixel can be regarded as two students $S^1_w$ and $S^2_w$, as shown in Fig. \ref{pixeltopixel}. The $S^1_w$ is generator as the main student. The $S^2_w$ is discriminator as the associate student.

The annotation of $S^1_w$ output comes from the $S^2_w$.
The label of $S^2_w$ output is defined by the source the input.

The cGAN loss is defined as follows,
\begin{equation}
\mathcal{L}_{cgan}(S^1_w, S^2_w; x, y) = \log(S^2_w(x, y)) + \log(1 - S^2_w(S^1_w(x), y)).
\label{cganloss}
\end{equation}
The $S^1_w$ and $S^2_w$ are related to cGAN loss. To optimize the parameters of $S^1_w$,  we minimize the cGAN loss when fixing the $S^2_w$.
From the original paper, PixeltoPixel uses $L_1$ distance rather than $L_2$ as $L_1$ encourages less blurring. 
\begin{equation}
\mathcal{L}_1(S^1_w(x), y) = \| S^1_w(x) - y \|
\label{L1dis}
\end{equation}
When we fix the parameters of $S^2_w$, $\mathcal{L}_{cgan}$ can be regarded as as a part of the loss of $S^1_w$.
The total loss for $S^1_w$ is
\begin{equation}
\mathcal{L}^1_{total}(S^1_w; x, y, S^2_w) = \mathcal{L}_{cgan}(S^1_w; x, y, S^2_w) + \lambda \mathcal{L}_{1}(S^1_w; x, y)
\label{pixeltopixelmain}
\end{equation}
where $\lambda$ is a hyperparameter to balance the two losses.

When we fix the $S^1_w$ to optimize the $S^2_w$, the total loss can be defined as a cross-entropy loss which is the same as the negative 
$ \mathcal{L}_{cgan}$.
\begin{equation}
\mathcal{L}^2_{total}(S^2_w; x, y ,S^1_w) = - \mathcal{L}_{cgan}(S^2_w; x, y, S^1_w).
\label{cross-entropy}
\end{equation}

The cooperative learning between $S^1_w$ and $S^2_w$ is that $S^1_w$ passes the generated images to $S^2_w$ and $S^2_w$ labels the outputs of $S^1_w$ then passes it back to $S^1_w$ through \eqref{cganloss}. With the cGAN's loss, $S^1_w$ and $S^2_w$ accomplish the data spreading and label spreading. Meanwhile, $S^1_w$ also conducts the individual learning through $\mathcal{L}_1$ loss.

\subsubsection{CycleGAN}
\label{cylceGAN}
\begin{figure}
\begin{center}
\includegraphics[scale=0.4]{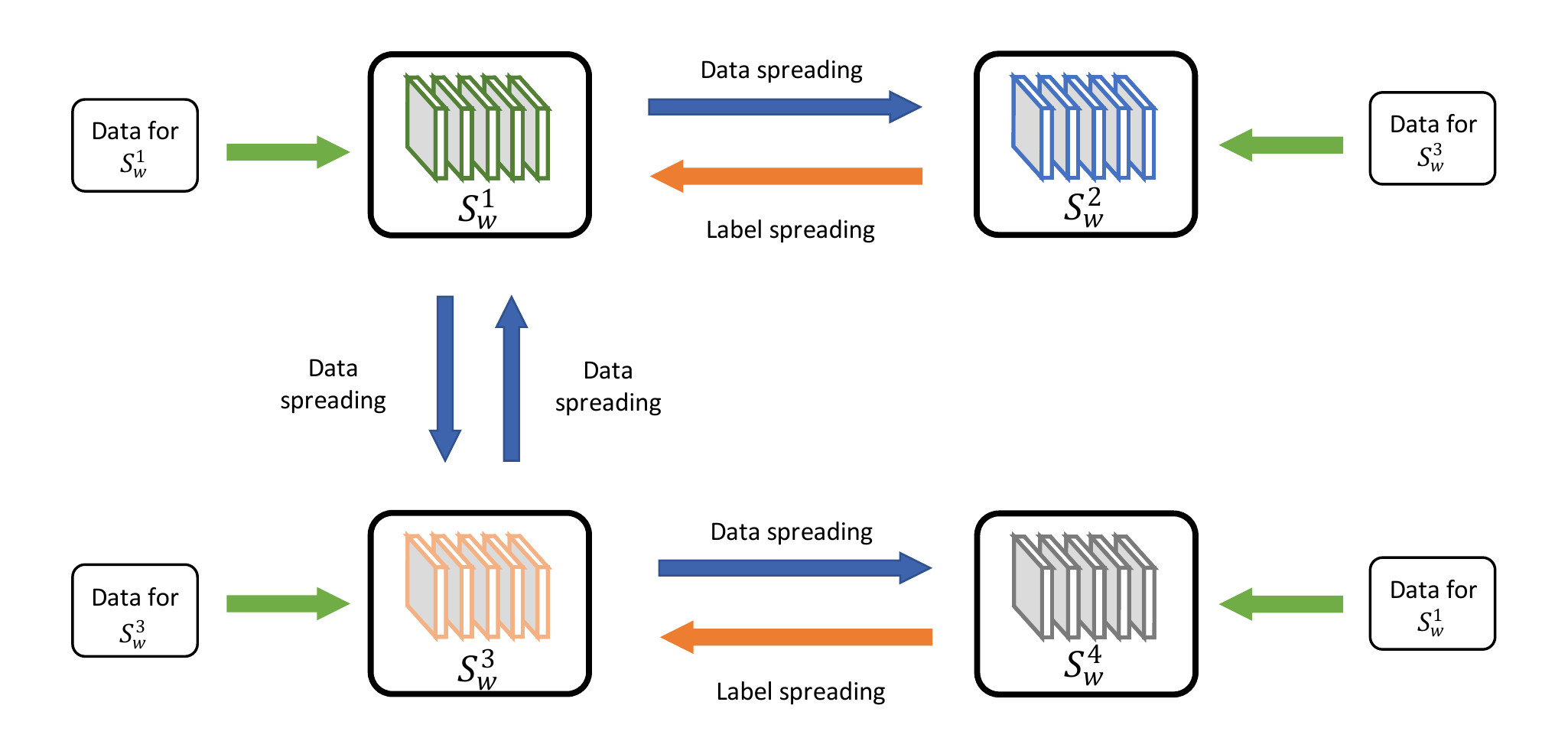}
\caption{Review the CycleGAN model based on IGC learning framework}
\label{cycleganmodel}
\end{center}
\end{figure}

CycleGAN is a general model to accomplish image to image translation on unaligned datasets. Suppose there are two image datasets $A$ and $B$, each of which contains a series of images about a specific object. The task is to replace the object in $A$ with the object in $B$, and vice versa. 
As is shown in Fig. \ref{cycleganmodel}, CycleGAN is composed of four student models $S^1_w$, $S^2_w$, $S^3_w$ and $S^4_w$. The $S^1_w$ and $S^2_w$ models are generator and discriminator respectively in cGAN.
The model $S^1_w$ achieves the transformation of images from $A$ to $B$. Similar to that, the model $ S^2_w $ judges whether the image comes from $A$.
The model $S^3_w$ achieves the transformation of images from $B$ to $A$.  The model $ S^4_w $ judges whether the image comes from $B$.  

The sample image in $A$ is denoted as $x_a$ and the sample image in $B$ is denoted as $x_b$.
The $S^1_w$ is related to $S^2_w$ and $S^3_w$.  The loss of $S^1_w$ related $S^2_w$ is called cGAN loss as follows,
\begin{equation}
\mathcal{L}^1_{cgan}(S^1_w, S^2_w; x_a, x_b) = \log(S^2_w(x_b)) + \log(1 - S^2_w(S^1_w(x_a)))
\end{equation}  
We defined the $z_a$ and $z_b$ for simplicity,
\begin{equation}
\begin{aligned}
z_a &= S^3_w(S^1_w(x_a))\\
z_b &= S^1_w(S^3_w(x_b))
\end{aligned}
\end{equation}
The loss of $S^1_w$ related $S^3_w$ is called the cycle consistency loss as follows,
\begin{equation}
\mathcal{L}^1_{cycle}(S^1_w; S^3_w, x_a) = L_p(z_a, x_a) =\left(\sum_{l=1}^{n}\left|z_a^{(l)}-x_a^{(l)}\right|^{p}\right)^{\frac{1}{p}}
\label{cycleloss}
\end{equation}
where $p$ sets the dimension of the metric space. When $p = 1$, the loss $\mathcal{L}_{cycle}$ is equivalent to the loss $\mathcal{L}_1$ in \eqref{L1dis}. When $p = 2$, the loss $\mathcal{L}_{cycle}$ is equivalent to the loss $\mathcal{L}_{pixel}$ in \eqref{L2dis}.

The total loss of the $S^1_w$ is summarized as follows,
\begin{equation}
\begin{aligned}
\mathcal{L}^1_{total} (S^1_w; S^2_w, S^3_w,x_a, x_b) = \ & \mathcal{L}^1_{cgan}(S^1_w; S^2_w, x_a, x_b) \\ + \ & \lambda\mathcal{L}^1_{cycle}(S^1_w; S^3_w, x_a)
\end{aligned}
\label{cycleL1}
\end{equation}

The total loss of the $S^2_w$ is similar to \eqref{cross-entropy}.
\begin{equation}
\mathcal{L}^2_{total}(S^2_w; S^1_w, x_a, x_b) = -\mathcal{L}^2_{cgan}(S^2_w; S^1_w, x_a,x_b).
\end{equation}
The total loss of the $S^3_w$ is similar to the loss of $S^1_w$ in \eqref{cycleL1}.
\begin{equation}
\begin{aligned}
\mathcal{L}^3_{total}(S^3_w; S^1_w, S^4_w, x_a, x_b)=\ & \mathcal{L}^3_{cgan}(S^3_w; S^4_w,x_a,x_b) \\ + \ & \lambda \mathcal{L}^3_{cycle}(S^3_w; S^1_w, x_b).
\end{aligned}
\end{equation}

\subsection{Guided Learning of style transfer}
\label{style_transfer}
Style transfer is the application of deep neural networks to transform a content image with a new style. 
A specific style is often defined by a style image.
The goal of the style transfer is to create a new image by combining a content image and a style image. The content image can be any picture, including landscape, portraits, etc., and the style image is required a certain style like in Fig. \ref{style_image}. The first method by Gatey \etal~\cite{7780634} is to accomplish the style transfer iteratively. 
This method is flexible enough to transform a content image with a variety of different styles. 
However, when a  series of content images need to be processed with a fixed style, the speed of the generating images becomes a bottleneck. Following the work of Johnson \etal~\cite{Johnson_2016}, they provide an online generative model to solve this problem. They train a convolutional network to accomplish a content image translation with a certain style. Compared to the Gatey \etal's method,  it produces similar qualitative results but is three orders of magnitude faster.
According to their contributions, more and more works follow up and develop in two directions \cite{Jing_2020}. 
Based on our proposed IGC learning framework, we believe that the difference between the two method lies in the student model.
Therefore, we introduce the iterative (online) generative model and the non-iterative (offline) generative model in a unified guided learning framework.

\subsubsection{Iterative generative model}
The method proposed by Gatey \etal~is a typical iterative generative method. The iterative generative method can be seen as an optimization-based method.
From the perspective of IGC, we categorize this method as guided learning and decompose it into a student model $S_w$ and a teacher model $T$ respectively.

A content image is denoted as $x_c$ and a style image is denoted as $x_s$.
Surprisingly, the $S_w$ is plain enough to be defined as a linear model.
\begin{equation}
S_{w} = w = x'
\end{equation}
where $w$ is initialized as white noise in accordance with the independent Gaussian distribution and $x'$ is the fused image combining a content image and a style image. 
There is no input to $S_w$ and the output is the optimized parameters for $S_w$.
The $T$ is a pretrained convolution neural network as a classifier model on ImageNet. Vgg network \cite{simonyan2014deep} was adopted as $T$ in \cite{7780634}, but other models like GoogleNet \cite{Szegedy_2015}  are still adequately appreciative. 

We first define the content loss for $x'$ and $x_{c}$  between $S_w$ and $T$ in guide learning.
\begin{equation}
\mathcal{L}^l_{c}(S_w; T, x_{c}) = \sum_{l}(T^l(x') - T^l(x_{c}))^2
\end{equation}
where $l$ means the $l$-th layer of the $T$ and $T^l(\cdot)$ denotes the activation of the $l$-th layer for inputs.
Before defining the style loss, we define the Gram matrix in the $k$-th layer of the $T$ for $x_s$.
\begin{equation}
G^k_{ij}(x_s)= \sum_{k}{T^k_{i}(x_c) T^k_{j}(x_c)}
\end{equation}
where $G^k_{ij}$ is the inner product between the vectorised feature maps $i$ and $j$ in the $k$-th layer.
Then, we define the style loss for $x'$ and $x_{style}$ as follows,
\begin{equation}
\mathcal{L}^k_{s}( S_w; T,  x_{s}) = \sum_{k}(G^k_{ij}(x') - G^k_{ij}(x_s))
\label{style loss}
\end{equation}
Finally, the total loss is defined for $S_w$.
\begin{equation}
\mathcal{L}_{total}{(S_w;T,x_c, x_s)} = \sum_{l \in L}\mathcal{L}^l_{c}(S_w; T, x_c) + \alpha \sum_{k \in K }\mathcal{L}^k_s{(S_w; T, x_s)}
\label{transferloss}
\end{equation}
where $L$ is a set of $l$, $K$ is a set of $k$ and $\alpha$ is a balance hyperparameter.

\subsubsection{Non-iterative generative model}
As the iterative generative model with a specific style picture, student model $S_w$ needs to be iteratively trained for each new content image. Another effective way is that the $S_w$ is pretrained for a certain style. When solving a new content image, the fused image $x'$ can be directly generated through $S_w$. For this purpose, Johnson \etal~\cite{Johnson_2016} replace the linear $S_w$ with a neural network and pretrain it to accomplish a non-iterative generative method. 

\begin{figure*}[t]
\begin{center}
\includegraphics[scale=0.6]{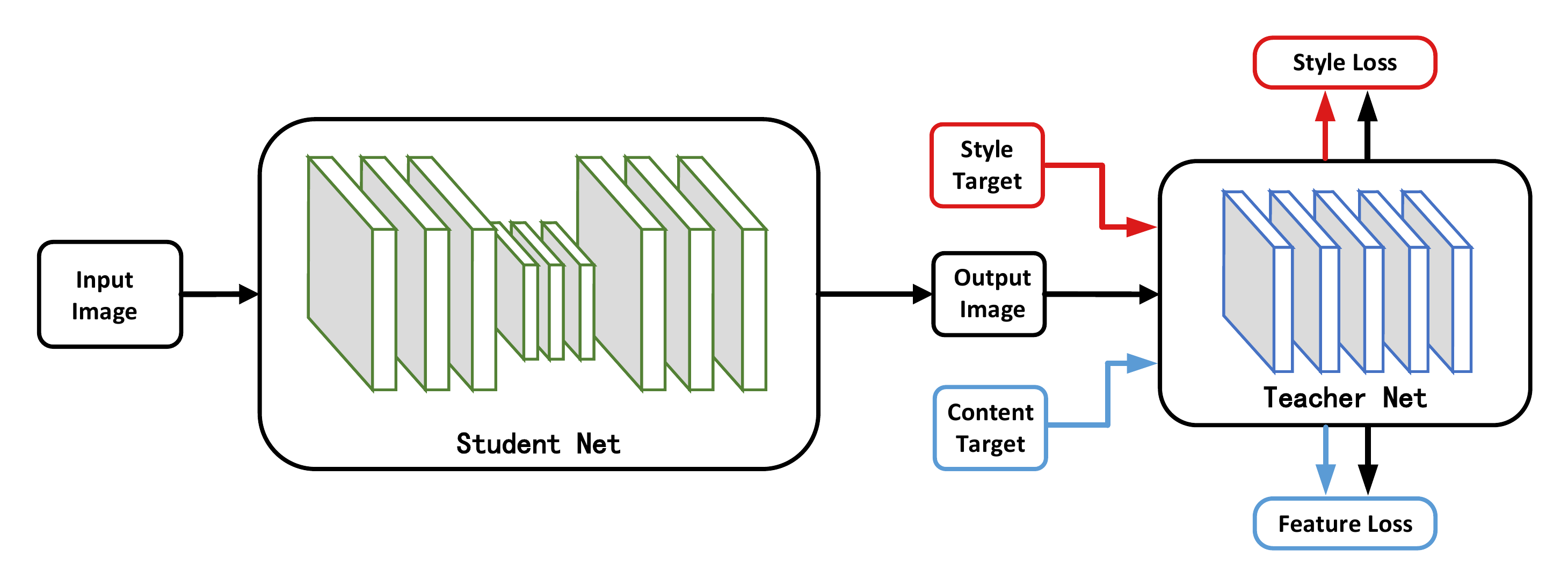}
\caption{The model of an off-line generative model on style transfer. The figure is modified from \cite{Johnson_2016} to fit the IGC learning framework.}
\label{liff}
\end{center}
\end{figure*}

The replaced $S_w$ has a similar structure in Fig. \ref{semantic}.
As is shown in Fig. \ref{liff}, we redrawn its model diagram based on IGC learning framework. By training an image dataset composed of content images, student model $S_w$ about the offline generative model can complete a certain style of end-to-end image transformation.

The loss is also similar to \eqref{transferloss}. They define perceptual loss functions that measure perceptual differences in content and style between images. 
Compared with the iterative generative model, the non-iterative generative model is only different in the structure and training method of the student model from the perspective of the IGC learning framework. 

\section{A Taxonomy of image transformation Methods}
\label{more_methods}
There are many methods about image transformation from different perspectives. 
According to the IGC learning framework in Table \ref{table_methods}, we roughly categorize the general image transformation methods  from two domains to multiple domains.
Correspondingly, we also divide the image style transfer works into multiple-style and arbitrary-style methods. 
The research of general image transformation and style transfer are developed independently. Wu \etal~\cite{8195348} presented a survey of image transformation about synthesis and editing with GANs, ignored the style transfer yet. Jing \etal~\cite{Jing_2020} investigated style transfer in recent years but did not cover general image transformation. 

In this article,  we review the general image transformation and style transfer along with a similar research trend.
To the best of our knowledge, we first organize the researches on general image transformation and style transfer from the unified perspective of our IGC learning framework. 
The development of general image transformation has a strong relationship with the development of style transfer. 
According to above discussion, we summary some methods of general image transformation and style transfer in Table~\ref{table_methods}.
First, we briefly introduce the problem to be solved by each method.
Then we discuss their models and learning methods under the IGC learning framework. More details about each method can refer to the original paper.

\begin{table*}
\begin{center}
\caption{The taxonomy of image transformation methods}
\begin{tabularx}{\textwidth}{lXXXX}
\hline
Methods &  Dataset Type & Learning Paradigm &  Loss & Description\\
\hline
FCN \cite{Shelhamer2017FullyCN} & Aligned & Independent learning & $\mathcal{L}_{pixel}$ & Image semantic segmentation. \\
U-Net \cite{u_net} & Aligned & Independent learning & $\mathcal{L}_{pixel}$ & Image semantic segmentation. \\
\hline
PixeltoPixel \cite{Isola_2017} & Aligned & Cooperative learning \& Independent learning & $\mathcal{L}_{pixel}$ \& $\mathcal{L}_{cgan}$ & General image translation. \\
TextureGAN \cite{Xian_2018} & Aligned & Cooperative learning \& Independent learning & $\mathcal{L}_{pixel}$ \& $\mathcal{L}_{cgan}$ \& Local texture loss & General image translation, especially in texture. \\
VidtoVid \cite{wang2018videotovideo} & Aligned & Cooperative learning \& Independent learning & $\mathcal{L}_{cgan}$ \& $\mathcal{L}_{style}$ (feature matching loss)  & Applying the video-to-video synthesis. \\
\hline
CycleGAN \cite{liu2017unsupervised} & Unaligned & Cooperative learning & $\mathcal{L}_{pixel}$ \& $\mathcal{L}_1$ \& $\mathcal{L}_{cgan}$ \& $\mathcal{L}_{cycle}$ & General image translation. \\
DiscoGAN \cite{kim2017learning} & Unaligned & Cooperative learning & $\mathcal{L}_{pixel}$ \& $\mathcal{L}_{gan}$ \& $\mathcal{L}_{cycle}$ & Similar to CycleGAN. \\
DualGAN \cite{Yi_2017} & Unaligned & Cooperative learning & $\mathcal{L}_{pixel}$ \& $\mathcal{L}_{cgan}$ \& $\mathcal{L}_{cycle}$ & Applying dual learning to image translation. \\
DA-GAN \cite{Ma_2018} & Unaligned & Cooperative learning & $\mathcal{L}_{gan}$ \& $\mathcal{L}_{cycle}$ \& Attention loss &  Apply the attention to two domains transformation. \\
\hline
UNIT \cite{liu2017unsupervised} &  Unaligned & Cooperative learning & $\mathcal{L}_{gan}$ \& $\mathcal{L}_{cycle}$ & Two domains transformation from probability distributions. \\
MUNIT \cite{Huang_2018} & Unaligned & Cooperative leaning & $\mathcal{L}_{cgan}$ \& $\mathcal{L}_{cycle}$ & Multiple domains transformation. \\ 
StarGAN \cite{Choi_2018} & Unaligned & Cooperative learning & $\mathcal{L}_{cgan} $ \& $\mathcal{L}_{cycle}$ & Multiple domains transformation, especially in face. \\
AugCGAN \cite{almahairi2018augmented} & Unaligned & Cooperative learning & $\mathcal{L}_{cgan}$ \& $\mathcal{L}_{cycle}$ & Extendingt the CycleGAN to multiple domains transformation. \\
\hline
Gatys \etal~\cite{Gatys2016Image} & Two images & Guided learning & $\mathcal{L}_{content}$ \& $\mathcal{L}_{style}$ & Iterative method to transfer the content image with any styles.\\
Johnson \etal~\cite{Johnson_2016}  & Content images and style images dataset & Guided learning & $\mathcal{L}_{content}$ \& $\mathcal{L}_{style}$ & Non-iterative method to accomplish a special style transformation with a network, one-style method.\\
Ulyanov \etal~\cite{Ulyanov_2017} & Content images and style images dataset & Guided learning & $\mathcal{L}_{content}$ \& $\mathcal{L}_{style}$ & Propose the instance normalization to improve the performance, one-style method.\\

StyleBank \cite{Chen_2017} & Content images dataset and some style images & Guided learning & $\mathcal{L}_{content}$ \& $\mathcal{L}_{style}$ & Propose filter banks for multiple styles. Every filter bank should be trained for one style, multi-style method.\\
Dumoulin \etal~\cite{dumoulin2016learned} & Content images dataset and some style images & Guided learning & $\mathcal{L}_{content}$ \& $\mathcal{L}_{style}$ & Propose a conditional instance normalization (CIN), multi-style methods. \\
\hline
Roy \etal~\cite{roy2019unsupervised}  & Content images and style images dataset & Guided learning & $\mathcal{L}_{content}$ \& $\mathcal{L}_{style}$ & Propose to learn a style network to adapt the affine parameters of CIN , arbitrary-style method. \\
AdaIN \cite{Huang_2017} & Content images and style images dataset & Guided learning  & $\mathcal{L}_{content}$ \& $\mathcal{L}_{style}$ & Propose to learn the channel-wise mean and variance feature statistics in activations to adapt the affine parameters of CIN , arbitrary-style method.\\
DIN \cite{roy2019unsupervised} & Content images and style images dataset & Guided learning & $\mathcal{L}_{content}$ \& $\mathcal{L}_{style}$ & The idea is similar to the Ghiasi \etal~but it improves the speed of style transfer, arbitrary-style method.\\ 
WCT \cite{roy2019unsupervised} & Content images and style images dataset & Guided learning & $\mathcal{L}_{content}$ \& $\mathcal{L}_{style}$ & Demonstrates that the covariance of the feature as the style reflects the whitening and coloring transformations, arbitrary-style method.\\
Lu \etal~\cite{Lu_2019} & Content images and style images dataset & Guided learning & $\mathcal{L}_{content}$ \& $\mathcal{L}_{style}$ & Derive a closed-form solution named Optimal Style Transfer (OST) and give a theoretical analysis in style transfer, arbitrary-style method.\\
\hline
\label{table_methods}
\end{tabularx}
\end{center}
\end{table*}

\subsection{Methods of General Image Transformation}
As mentioned in Section \ref{general} and Section \ref{cooperation learning},  most image transformation tasks focus on the objects in images. PixeltoPixel implements pixel-level image-to-image transformation on aligned datasets while CycleGAN implements object-level image transformation on unaligned datasets. Both methods use the GAN as the basic architecture, but the learning methods are different. From the view of the IGC learning framework, in PixeltoPixel, the tasks of $S^1_w$ and $S^2_w$ are different, where $S^1_w$ is the main task to implement image transformation and $S^2_w$ is the associate task to determine whether the image is forged. 
CycleGAN can be regarded as a model with two PixeltoPixels, the $S^1_w$
and $S^2_w$ as one, $S^3_w$ and $S^4_w$ as the other.
The $S^1_w$ and $ S^3_w$ are mutually dual models to realize the transformation from A to B and B to A respectively. The cycle loss as the pixel loss between $S^1_w$ and $S^3_w$ is adopted to train their parameters on unaligned datasets. 

However, both PixeltoPixel and CycleGAN have limited scalability, which means that one model only solves the image translation in two domains. The general image transformation among $N$ domains requires $N* (N-1)$ models.
To improve the flexibility and diversity, some researchers pay attention to accomplishing general image transformation in multiple domains with single model. 
In this section, we first discuss the general image transformation in two domains. Then, the methods of multiple domains transformation are introduced.

\subsubsection{Two domains transformation}
At the beginning, semantic segmentation is widely studied as a task to comprehend the content of images for decades.
The researchers expanded  the problem to general image transformation in two domains.
According to whether it is applicable to the aligned and unaligned datasets, 
we divide the model into two parts, one part is about individual learning, the other part is about cooperative learning.

For aligned datasets, most methods have similar structures as shown in Fig. \ref{semantic}. $\mathcal{L}_1$ and $\mathcal{L}_{style}$ loss are also commonly used.
Some methods improve the PixeltoPixel  from different aspects. TextureGAN \cite{Xian_2018} investigates the image synthesis guided by texture and develop a local texture loss to generate more realistic images. 
Studying the loss function on images can improve the quality of image translation under individual learning. 
Wang \etal~\cite{wang2018videotovideo} extend the image-to-image synthesis problem and investigate the potential research of the video-to-video synthesis. The most critical thing in video translation is the speed of image-to-image translation. The efficiency of image-to-image translation is also a problem that neural network researchers focus on.
However, these methods on the alignment datasets have the same problem, i.e., it is difficult to extend to general image transformations due to the cost of labelling datasets.  

For unaligned datasets, learning from unlabelled data is the trend of deep image research. It is difficult for the aforementioned CycleGAN to clearly confirm whether it is the first work.
The idea of CycleGAN about cycle loss was also proposed by other models in the same period, e.g., DiscoGAN \cite{kim2017learning} and DualGAN \cite{Yi_2017}.  DiscoGAN and DualGAN both solve the general image transformation in two domains and have similar architectures which related to the four student models like CycleGAN. The difference is that the details of student model and experiments on the image-to-image translation. The experiment of DiscoGAN verified the necessity of cycle loss that ensures the correlation between domain A and domain B. Inspired by dual learning from natural language translation \cite{10.5555/3157096.3157188}, DualGAN developed a novel dual-GAN mechanism to accomplish image-to-image translation between two domains. The DualGAN also utilizes Wasserstein GAN \cite{gulrajani2017improved} to improve the stability of training and the results of experiments. 
To improve the quality of general image transformation, some researchers have proposed new losses under cooperative learning.
The unsupervised image-to-image translation (UNIT) \cite{liu2017unsupervised} learns the marginal distributions to infer the joint distribution of images in different domains. DA-GAN \cite{Ma_2018} learns the instance-level correspondences by attention mechanism \cite{vaswani2017attention}.

\begin{figure}[t]
\begin{center}
\includegraphics[scale=0.35]{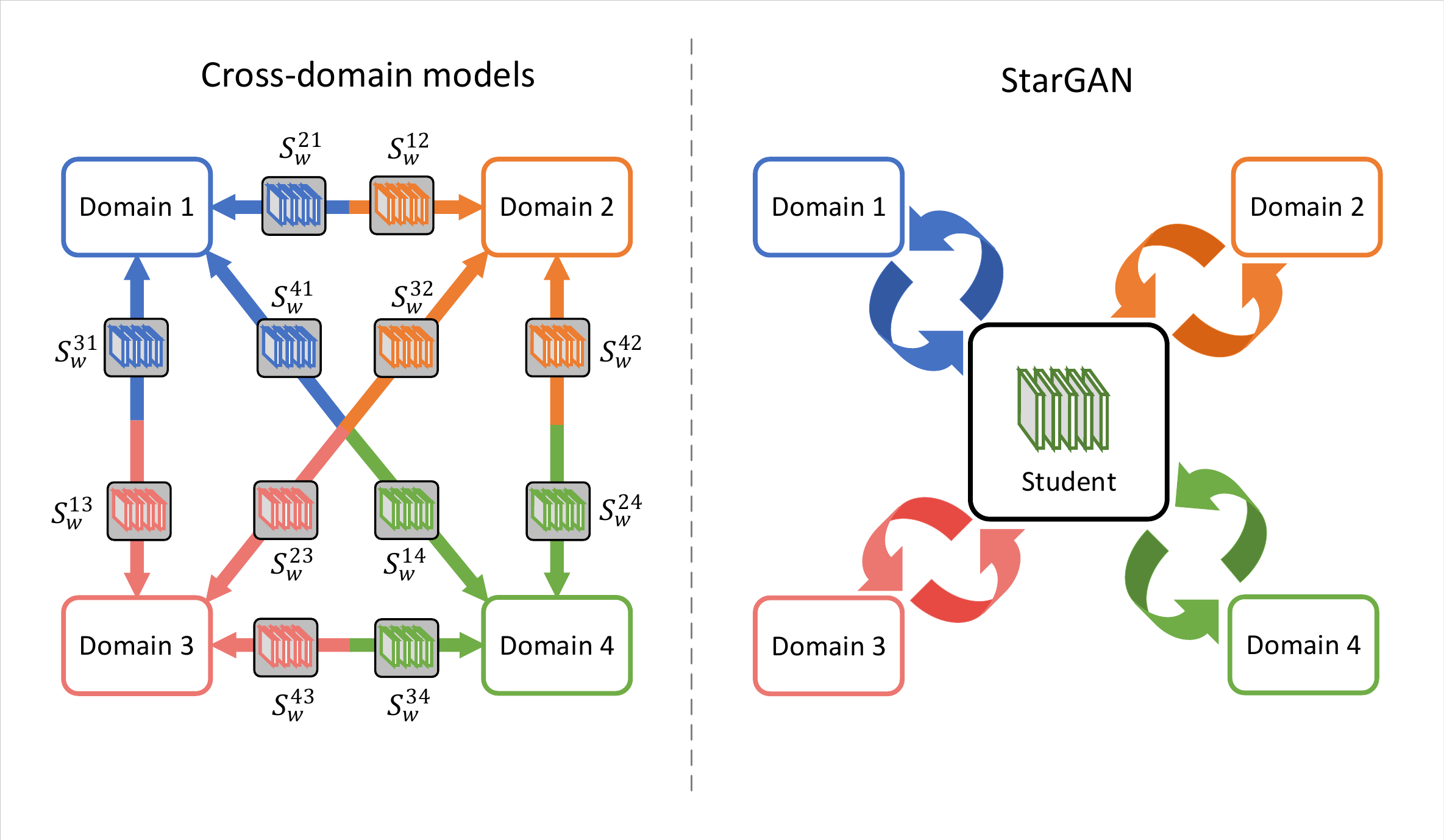}
\caption{Comparison between models about two domains and StarGAN about multiple domains \cite{Choi_2018}. (a) To handle multiple domains, cross-domain models should be built for every pair of image domains. (b) StarGAN has capable of learning mappings among multiple domains using a single generator. }
\label{stargan}
\end{center}
\end{figure}

\subsubsection{Multiple domains transformation}
Multiple domains transformation aims to convert images with multiple types in only one model.
By solving the poor flexibility on the two domains transformation,
the methods about multiple domains transformation have received extensive attentions and researches.  Inspired by cGAN, most methods try to adapt to the migration of images on multiple domains by conditional input. 

\begin{figure}[t]
\begin{center}
\includegraphics[scale=0.40]{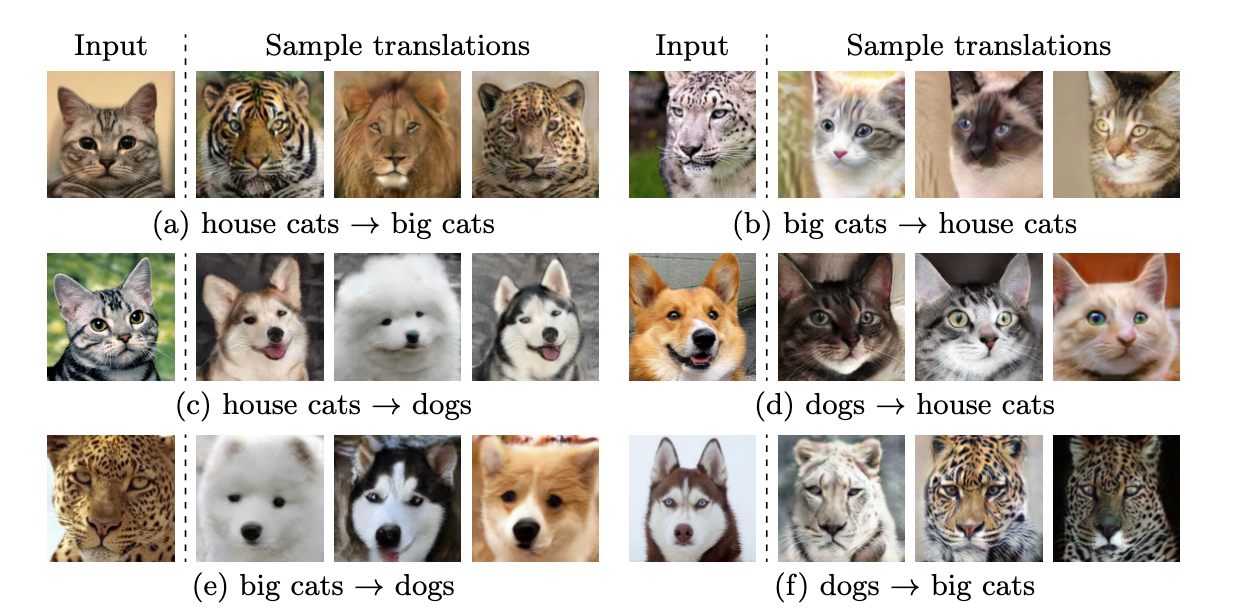}
\caption{Examples of UNIT. It accomplished Image-to-image translation among similar objects. The figure comes from \cite{liu2017unsupervised}.}
\label{munit}
\end{center}
\end{figure}

\begin{figure}[t]
\begin{center}
\includegraphics[scale=0.40]{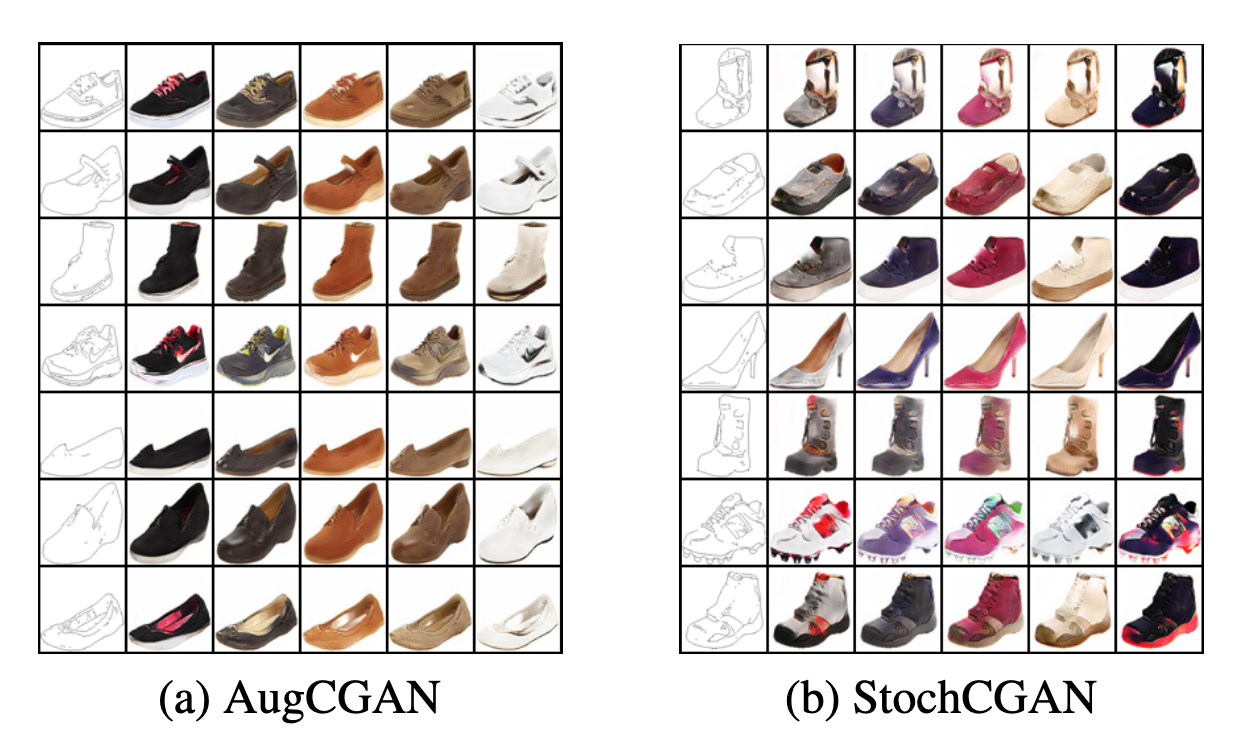}
\caption{Example results of Augmented CycleGAN. StochcGAN means the stochastic CycleGAN as the comparative method. The figure is credited from~\cite{almahairi2018augmented}}
\label{shoes}
\end{center}
\end{figure}

However, due to the lack of annotation information, the quality of image transformation on cooperative learning is not as good as that on individual learning. There are constraints and connections among multiple domains.
StarGAN \cite{Choi_2018} aims to achieve transformation among multiple domains of facial attributes and expressions.
Fig. \ref{stargan} shows the two domains transformation method and StarGAN. The model of StarGAN determines the image translation domain according to the conditional input. 
It utilizes a mask vector as the conditional input to control all available domain labels, and then achieves the mapping between arbitrary domains with a single model. Improved from UNIT, Multimodel Unsupervised Image-to-Image Translation (MUNIT) \cite{Huang_2018} applies one model to infer the joint probabilities among multiple image domains. 
MUNIT assumes that the image representation can be decomposed into a content code that is domain-invariant and a style code that captures domain-specific properties. In the considerations of the learning \cite{locatello2018challenging}, MUNIT explicitly maps the image to a content and a style and accomplishes image-to-image translation in multiple domains by combining different styles with special content. MUNIT implements image transformation among multiple similar objects in Fig \ref{munit}. Augmented CycleGAN (AugCGAN) \cite{almahairi2018augmented}  extends CycleGAN to many-to-many mappings among image domains. AugCGAN adds a latent variable to the original image as input and also outputs a new latent variable with the translation image. As shown in  Fig. \ref{shoes}, multiple translation images can be generated from the same source image with additional information. It successfully accomplishes the diversity of image translation. 
We summary two main objectives in multiple domains transformation. One is to convert the attributes about object, the other is to transfer among similar objects.



\subsection{Methods of Style Transfer}
\label{styletransfer}
Most methods about style transfer are related to the guided learning in the IGC learning framework. Just like the development of general image transformation, we survey the methods of style transfer according to similar classification. 
When the content image is appointed, iterative methods like Gaty \etal~can realize the transformation of any styles. However, due to the efficiency of iterative methods, most proposed methods are non-iterative methods. For example, the student model need to pretrained on a special style in Johnson \etal~
A simple method of multiple-style transfer is to pretrained multiple student models of different styles. 
The adaptability and efficiency of style transfer methods are crucial for their applications yet. 
The problem is similar to the aforementioned general image transformation among multiple domains.
Many follow-up researchers study non-iterative methods to achieve the transfer about multiple styles, even arbitrary styles. 

Before discussing those methods about non-iterative methods, we introduce important developments in iterative generative methods. Many subsequent studies are also based on those works. 
Instance normalization \cite{ulyanov2016instance} is proposed instead of batch normalization \cite{ioffe2015batch}. Instance normalization, which is equivalent to batch normalisation when the batch size is set to $1$,  leads to a significant improvement in stylisation quality \cite{Ulyanov_2017}. Instance normalization methods rely on affine transformations to produce arbitrary image style transfers, of which the parameters are computed in a pre-defined way. Dynamic Instance Normalization (DIN) \cite{Jing_2020} introduces a sophisticated style encoder to automatically learn the affine parameters of transformations to provide flexible support for image style transfer. The iterative generation method can generally be regarded as an arbitrary style method. 

As for non-iterative methods, most methods focus on improving student model from the respective of IGC learning framework.
For multiple styles, one of the main purposes is to use a single student model to transfer multiple styles according to the addition input conditions. For arbitrary styles, a student model can accomplish content image  transformation with arbitrary style image, just like iterative methods. 
Since most solutions try to modify the student model, we mainly introduce the student model of each methods. The teacher model and learning methods are similar to the work of Johnson \etal~from the respective of IGC learning framework. 
The student model of style transfer is also based on the encoder and decoder.
Most methods propose a style network model that extracts the style from the image and embed it between the encoder and the decoder. In this Section, we mainly focus on the no-iterative generative methods and also summary them in Table \ref{table_methods}.

\subsubsection{Multiple-Style methods}
\begin{figure*}
\begin{center}
\includegraphics[scale=0.5]{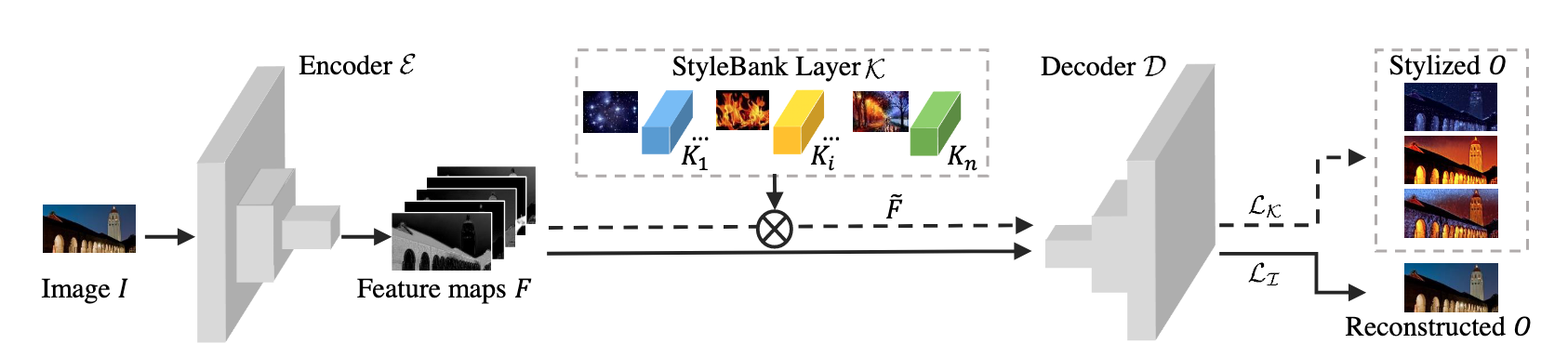}
\caption{StyleGAN consists of three modules: image encoder $\mathcal{E}$, StyleBank layer $\mathcal{K}$ and image decoder $\mathcal{D}$. The figure comes from \cite{Chen_2017}}
\label{styleganmodel}
\end{center}
\end{figure*}

To handle multiple styles with a single student model, a way is to separate image content and combine them when a special style is appointed.
StyleBank \cite{Chen_2017} is composed of multiple convolution filter banks and each filter bank explicitly represents one style.
StyleBank explicitly divides the student model into encoder, style filter bank, and decoder. To transfer an image to a specific style, the corresponding filter bank is operated on top of the intermediate feature embedding produced by a single encoder. StyleBank realizes multiple styles from content image to fusion image by defining different style bank filters in Fig. \ref{styleganmodel}. 
Although StyleBank avoids training a whole model, it still needs to train special filter bank for each style.

Inspired by instance normalization, Dumoulin \etal~\cite{dumoulin2016learned} propose a conditional instance normalization (CIN) that learns a different set of parameters $\alpha_s$ and $\beta_s$ for each style $s$. The conditional instance normalization is defined as,
\begin{equation}
{CIN}_{\mathcal{F}_i(x_c)} = \alpha_s\frac{(\mathcal{F}_i(x_c) - \mu(\mathcal{F}_i(x_c))}{\sigma(\mathcal{F}_i(x_c))} + \beta_s,
\label{CIN}
\end{equation}
where $\mathcal{F} $ is the activation of the $i$-th feature layer of the student model. 
They found that different styles are accomplished by using the same convolution parameters but different affine parameters in instance normalization layers. This critical discovery makes it possible to learn multiple styles in a network with stacked CINs.

\begin{figure*}[h]
\begin{center}
\subfigure[Semantic Mask]{
\includegraphics[scale=0.4]{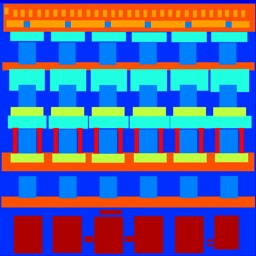}
\label{result1}
}
\subfigure[Ground Truth]{
\includegraphics[scale=0.4]{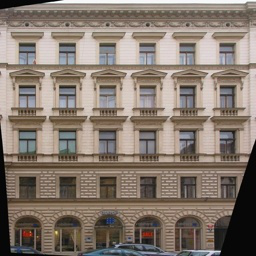}
\label{groundtruth}
}
\subfigure[By U-Net]{
\includegraphics[scale=0.4]{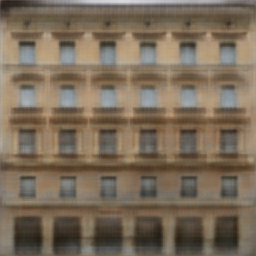}
\label{unet}
}
\subfigure[By Pixel2Pixel]{
\includegraphics[scale=0.4]{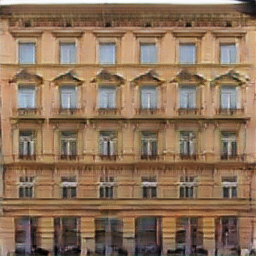}
\label{pix2pix}
}
\end{center}
\caption{Some example results of image-to-image translation generated by U-Net in Fig.~\ref{unet} and Pixel2Pixel in Fig.~\ref{pix2pix}. The ground truth is in Fig.~ \ref{groundtruth}.}
\label{vis1_build}
\end{figure*}

\subsubsection{Arbitrary-Style methods}
The style of any image is extracted through the deep convolution network, and as a part of a student model, arbitrary styles are synthesized on content image.
Following conditional instance normalization, Ghiasi \etal~\cite{ghiasi2017exploring} train a style transfer network to supply a set of affine parameters.
Since its style transfer network is trained on corpus of style images and content images, it has a certain generalization to extract styles for unknown images. 
Besides, AdaIN \cite{huang2017adain} also solves a similar problem.  Instead of training an entire style network, AdaIN proposed adaptive instance normalization layer embedded in the original autoencoder. 
Modifying the \eqref{CIN}, the adaptive instance normalization is defined as,
 \begin{equation}
 \begin{array}{l}
 AdaIN_{\mathcal{F}_i}(x_c, x_s) = \\
 \quad \quad \sigma(\mathcal{F}_i(x_s))\frac{(\mathcal{F}_i(x_c) - \mu(\mathcal{F}_i(x_c))}{\sigma(\mathcal{F}_i(x_c))} + \mu(\mathcal{F}_i(x_s)).
 \end{array}
\label{AdaIN}
\end{equation}
AdaIN transfers the channel-wise mean and variance feature statistics between content and style feature activations.  
AdaIN does not learn fixed affine parameters, but calculates them adaptively from the pattern image.
Drawing on the advantages of AdaIN and Ghiasi \etal~' method, Dynamic Instance Normalization (DIN) \cite{din} introduces a sophisticated style encoder and a compact lightweight content encoder for fast inference.  DIN uses a lightweight architecture to achieve arbitrary style transfer, which reduces the computational cost by more than $20$ times.

However, only adjusting the mean and variance of feature statistics makes it hard to synthesise complicated style patterns with rich details and local structures. Whitening and Colouring Transform (WCT) \cite{roy2019unsupervised} demonstrates that  whitening and colouring transformations reflect a direct matching of feature covariance of the content image to a given style image. 
WCT adopts feature transformation network to accomplish whitening and colouring, which is embedded to decoder as the style for image composition. Lu \etal~\cite{Lu_2019} give a unique closed-form solution to image style transfer as an optimal transport problem.
They also demonstrate the relations of their formulation with former works like AdaIN and WCT. 
Both AdaIN and WCT holistically calculate their style feature distributions and the parametrize feature statistics. 
Avatar-Net \cite{shen2020scale} proposed zero-shot artistic style transfer that means the same as transferring arbitrary style into content images. 
The key ingredient of avatar-Net is a style decorator that makes up the content features by semantically aligning style features from an arbitrary style image. Avatar-Net does not only holistically match their feature distributions but also preserves detailed style patterns in the decorated features.

\section{Experiment}
\label{experiment}
In this section, we perform experiments about image transformation on three learning paradigms of the IGC learning framework. We select several models on three types of datasets from the IGC learning framework. We qualitatively evaluate each model through metrics and visualization to show the superiority in context of the IGC learning.

\subsection{General Image Transformation with Independent Learning}
\label{experment1}
In Table \ref{dataset}, we choose one aligned datasets: CMP Facades. We choose the U-Net as a independent learning model and PixeltoPixel as a cooperative learning model.
The codes \footnote{https://github.com/junyanz/pytorch-CycleGAN-and-pix2pix}
come from the open source library by official.
\subsubsection{Experiment details}
The CMP Facades consists of $606$ images. 
Each image is cropped as 256 x 256 x 3 as the input. The dataset is divided into training, validation, and testing datasets with $400$, $100$, and $106$ images respectively.
U-Net is composed of $4$ skip connection blocks like in Fig \ref{semantic}. Batch normalization \cite{ioffe2015batch} is adopted instead of dropout layer \cite{JMLR:v15:srivastava14a} in each blocks. U-Net as a student model is trained by \eqref{L1dis}. Taking U-Net as the main student, PixeltoPixel adds PatchGAN \cite{li2016precomputed} as an associate student. PatchGAN encourages U-Net to generate greater color diversity but has no effect on spatial statistics by classifying whether a pixel is real or not. In PixeltoPixel, the U-Net and PatchGAN is trained by \eqref{pixeltopixelmain} and \eqref{cross-entropy} respectively. 
We set $\lambda = 10$ in \eqref{pixeltopixelmain} and use the Adam solver \cite{kingma2014adam} with a batch size of $1$.
All networks are trained from scratch with a learning rate of $0.0002$.
We also decay linearly the learning rate of each parameter group by $\gamma = 0.1$ with every $50$ epochs.
U-Net and  PixeltoPixel are trained totally for $200$ epochs on training dataset.

\subsubsection{Evaluation metrics}
This task is to transform a label image into a realistic image. 
Evaluating the quality of synthesized images is an open and difficult problem. We will discuss more about the evaluation of image transformation in Section \ref{evaluation_problem}. 
In this experiment, the per-pixel loss in \eqref{L2dis} can be seen as the metric to access the quality between the generated image and ground truth. However, it cannot measure the realism of the image. Therefore, we visualize some transistorized images for intuitive comparison. 

\subsubsection{Visualization and analysis}
Table \ref{losstable1} shows the mean $L_1$ loss in \eqref{L1dis} and per-pixel loss in \eqref{L2dis} of these models on testing dataset. With distance-based loss, the difference between the generated image and the ground truth is small. We can not judge the quality of the images from the loss. As is shown in Fig. \ref{vis1_build}, we visualize some generated images by different models with the semantic label and ground truth. From Fig. \ref{vis1_build},  the generated images by U-Net is blurred and can easily be distinguished among those images. The images generated by PixeltoPixel have more details, making it difficult to distinguish from the ground truth. From the perspective of the IGC learning, PatchGAN as an assistant student does provide some annotation for the main student by cooperative learning. Visual analysis is more in line with human intuition, while distance-based loss is more difficult to evaluate the performance of image transformation methods. 

\begin{table}[h]
\begin{center}
\caption{The mean $L_1$ loss for U-Net and PixeltoPixel models.}
\begin{tabular}{ccc}
\hline
Models & $L_1$ Loss & Per-pixel Loss\\
\hline
U-Net & 0.4891 & 0.3699 \\ 
PixeltoPixel & 0.4921 & 0.3713 \\
\hline
\label{losstable1}
\end{tabular}
\end{center}
\end{table}

\begin{figure*}
\begin{center}
\subfigure[Original Image]{
\includegraphics[scale=0.4]{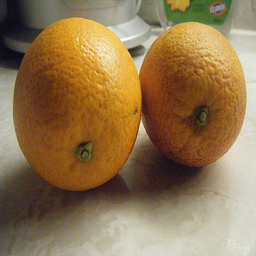}
\label{original}
}
\subfigure[Object Transformation]{
\includegraphics[scale=0.4]{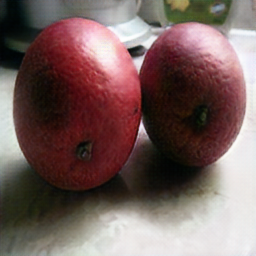}
\label{objecttran}
}
\subfigure[Original Image]{
\includegraphics[scale=0.4]{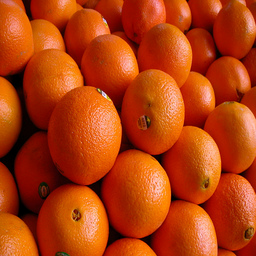}
\label{reconstruct}
}
\subfigure[Object Transformation]{
\includegraphics[scale=0.4]{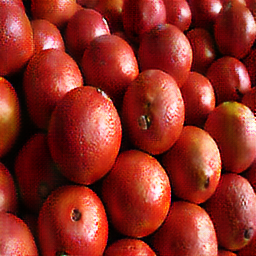}
\label{original}
}
\end{center}
\caption{Some example results of image transformation with cooperative learning generated by CycleGAN on appletoorange. }
\label{vis2}
\end{figure*}

\begin{figure*}
\begin{center}
\subfigure[G\_A and D\_A Losses]{
\includegraphics[scale=0.40]{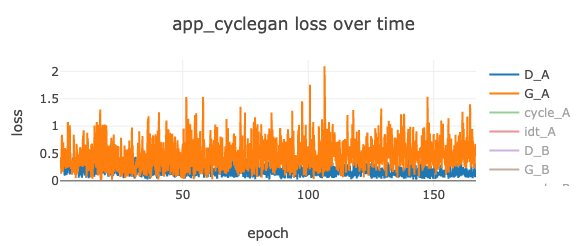}
\label{loss:loss1}
}
\subfigure[Cycle\_A and Idt\_A Losses]{
\includegraphics[scale=0.40]{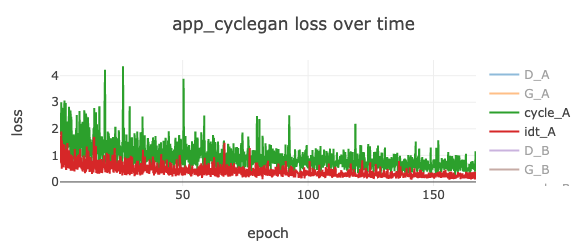}
\label{loss:loss2}
}
\end{center}
\caption{The loss of CycleGAN when trained on appletoorange dataset. G\_A demonstrates the $\mathcal{L}^1_{cgan}$ of $S^1_w$ and D\_A shows the $\mathcal{L}^2_{cgan}$ in Fig. \ref{loss:loss1}.  As shown in Fig. \ref{loss:loss2}, cycle\_A means the $\mathcal{L}^1_{cycle}$ of $S^1_w$ in \eqref{cycleloss} and idt\_A represents the $\mathcal{L}^1_{identity}$ in \eqref{identity}.}
\label{losscycgan}
\end{figure*}

With the help of GAN on cooperative learning, the assistant model instructs the main model to generate more texture details. 
It is difficult to describe the details of texture by loss defined on the main model.
Therefore, cooperative learning is a good way to learn knowledge among models. 
Furthermore, since the main students of PixeltoPixel are the same as U-Net, the cost of image transformation is similar to each other. Cooperative learning in the training process does not harm the efficiency of the model in inference. Under the guidance of the IGC framework, we can study to apply multiple learning methods to the previous single learning method. Many works introduce GANs to improve the main model.  Combined with individual learning, cooperative learning can provide additional information to guide the main students' learning.

\subsection{General Image Transformation with Cooperative Learning}
In realistic image scenes, the collected data is often not aligned. For example, we can obtain a bunch of misaligned images through a deep classification network. In this section, we mainly experimented the general image-to-image translation on unaligned data under cooperative learning.
According to Table \ref{dataset}, we choose one unaligned datasets: apple2orange \footnote{ \url{https://people.eecs.berkeley.edu/~taesung_park/CycleGAN/datasets/apple2orange.zip}}. 
The dataset is collected by \cite{Zhu_2017}. We verify the experiment of the CycleGAN model and discuss its performance and shortcomings.

\subsubsection{Experiment details}
The apple2orange dataset contains $1261$ images of apples and $1267$ images of oranges. The images are crawled from the web with different resolutions, but uniformly cropped to $256$ x $256$ and normalized as the input of the CycleGAN.
The details of CycleGAN are discussed in Section \ref{cylceGAN}.  However, in the implementation, the identity loss is also defined for $S^1_w$ as following.
\begin{equation}
\mathcal{L}^1_{identity}(S^1_w;x_a)= \left| \left| S^1_w(x_a) - x_a \right| \right|_1
\label{identity}
\end{equation}
The identity loss encourages the mapping to preserve color composition between the input and output.
Besides, we also use adam with the batch size of $1$ and the learning rate of $0.0002$. We iteratively train the model for $200$ epochs. The losses related $S^1_w$ in the training is shown in Fig. \ref{losscycgan} and the losses of $S^3_w$ is similar to it. We can estimate the training of CycleGAN by monitor the loss in \ref{loss:loss2}.

\begin{figure*}
\begin{center}
\includegraphics[scale=0.62]{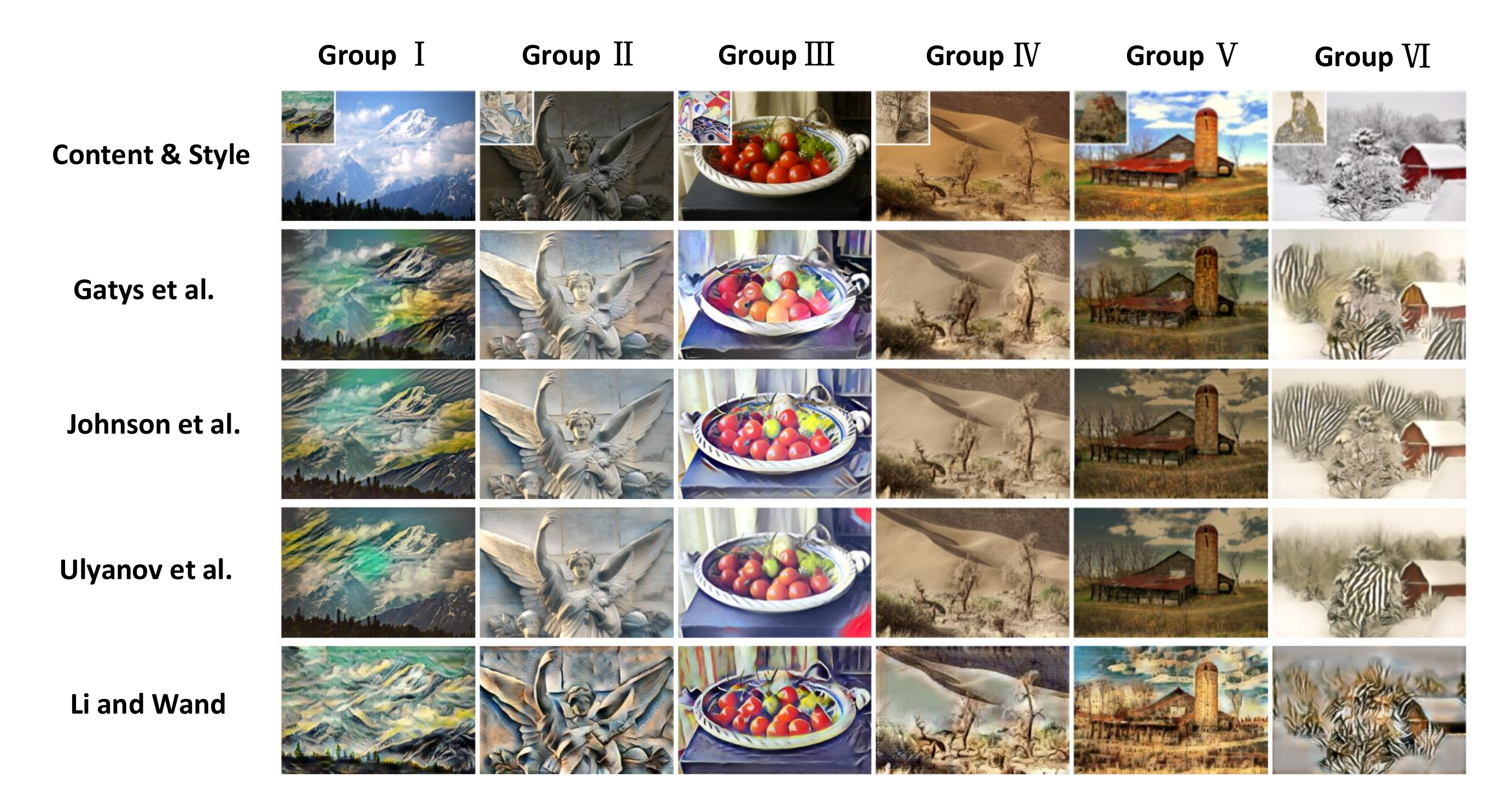}
\end{center}
\caption{Some samples of aforementioned methods about image style transfer for qualitative evaluation. The content images are from the benchmark dataset proposed by Mould and Rosin \cite{10.5555/2981324.2981327}. 
The figure comes from \cite{Jing_2020}.}
\label{vis3}
\end{figure*}

\subsubsection{Evaluation method}
For unaligned datasets, we cannot directly utilize the ground truth to evaluate the generated images. Evaluating the quality of generated images is a challenging problem, which we will discuss more in section \ref{evaluation_problem}. Following the previous methods, we visually analyse the results of CycleGAN as a typical cooperative learning in the unaligned dataset. 

\subsubsection{Visualization and analysis}
Fig. \ref{vis2} demonstrates some examples about the transformation from orange to apple for the trained CycleGAN. The transformation object in Fig. \ref {objecttran} is a typical example, which represents the result of most images. However, we think the transformation is not good enough to consider as an apple. The transformation occurs more in the color of the object, but the shape of the object changes less. This problem also exists but it was ignored in the original paper.
They experimented on datasets such as horses to zebras, which have highly similar shapes and different textures. 
From the IGC learning, we believe that they adopt a naive cycle consistency loss in \eqref{cycleloss}.  As is shown in Fig. \ref{reconstruct}, the cycle consistency restricts the object transfiguration to generate the blur images. The cycle consistency can be seen as the extension of $\mathcal{L}_1$ and  $\mathcal{L}_2$ on unaligned datasets. Some researchers utilize the new classification network as a teacher model to strengthen the modification of target shapes like face transformation \cite{perov2020deepfacelab}. The new methods of cooperative learning or guided learning are proposed to improve the performance of main student.

\subsection{Style Transfer with Guided Learning}
The survey of style transfer in \cite{Jing_2020} makes sufficient experiments to show the results of the related methods.  Therefore, we summary their works and review them with the IGC learning framework. The relationships among the models are discussed in this subsection.  

\subsubsection{Experiment details}
Jing \etal select and test ten style images. Some style transfer method is offlined and trained on MS-COCO dataset. Similar to what we classify in Section \ref{style_transfer}, they also divide style transfer methods into two categories: 1)Image-Optimisation-Based Online Image transformation (IOB), 2)Model-Optimisation-Based Offline Image transformation (MOB). The method proposed by  Gatys \etal~belongs to IOB and other methods proposed by Ulyanov \etal~\cite{ulyanov2016texture},  Johnson \etal~\cite{Johnson_2016} and Li \& Wand \cite{Li_2016} belong to MOB. 
The works of Gatys \etal~and Li \& Wand are introduced as the typical iterative generative and non-iterative generative methods in Section \ref{style_transfer}.  Ulyanov \etal focus on the synthesis of textures to improve the IOB on style transfer. Furthermore, following this work, they also improve their method by applying the instance normalization to maximize quality and diversity of style transformation in \cite{Ulyanov_2017}. Li \& Wand achieve the style transfer by precomputed real-time texture synthesis with markovian generative adversarial networks. From our IGC learning view, Ulyanov \etal~optimize the structure of the main student to accomplish the better style transfer and Li \& Wand \cite{Li_2016} introduce a adversarial network as a new assistant student to improve the style transfer network through cooperative learning.
We can evaluate image style in terms of the quality of generated images. 
 
\subsubsection{Visualization and discussion}
Fig.~\ref{vis3} shows the results of related methods. The IOB method is computationally expensive whose results are appealing in visually. The method of Gatys \etal~usually is regarded as the gold-standard method in the community of style transfer. For the MOB methods,  Ulyanov \etal~and Johnson \etal~have a similar idea but differ in their detailed network architectures. Li \& Wand use cGAN to transmit the image, which has some strange rendering, as shown in the middle area of the image in the group one of Fig.~\ref {vis3}. The student model of IOB method is too simple to complete the transfer of image styles in one step. It needs to learn the transfer style step by step through the guidance of the teacher model. The student models of MOB methods can learn under the guidance teacher model and independently accomplish the style transfer after training. The IOB and MOB methods are essentially the same in the student models.

\section{Application}
\label{application}

With the development of general image transformation and style transfer in neural network, the researchers also apply those technology to related projects. Some works create more complex tools to extend image processing. Those tools bring great convenience to image and video processors. In the field of image and video creation, we can generate the forgery image and video based on the general image transformation. Furthermore, some applications are prevalent by younger to provide social communication and entertainment. As following, we briefly introduce three aspects of application in image transformation. 

\begin{figure}[t]
\begin{center}
\includegraphics[scale=0.45]{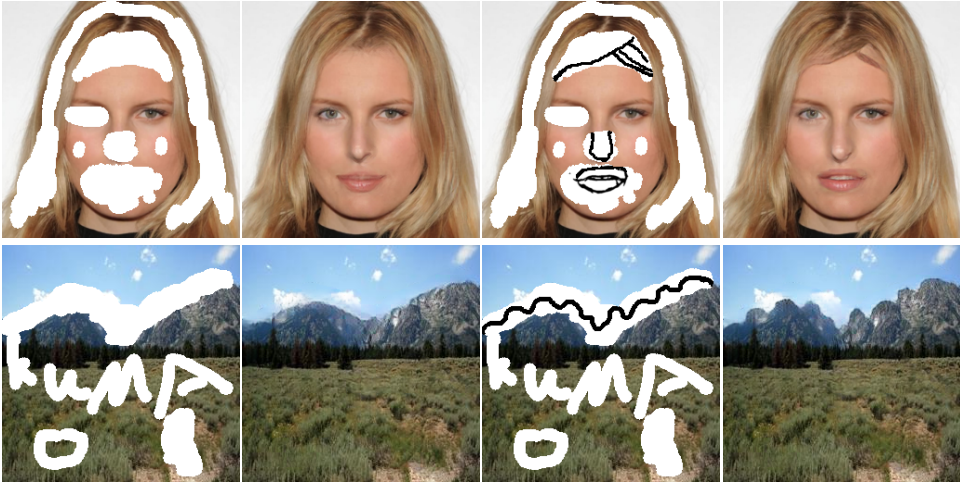}
\end{center}
\caption{Face editing on CelebA \cite{liu2018large}. The beard on the face has been edited in each row. The original figure is from \cite{chen2018isolating}.}
\label{image_editing}
\end{figure}

\subsection{Tools for Image Transformation Works}
image transformation have a broad application about various tasks of image restoration and manipulation. As aforementioned the general image  transformation methods in Section \ref{more_methods}, more and more researchers pay attention to applying them to actual application scenarios.

Image manipulation in CV is a little different from the image processing in graphic. Image manipulation deals with more complex tasks, which are usually difficult to define exactly by mathematics. Image inpainting \cite{Elharrouss_2019,9156579} and Super-resolution \cite{wang2020deep} are two important applications about image manipulation. Image inpainting is a task of reconstructing missing regions in an image.  For example, as is shown in Fig. \ref{vis1_build} in Section \ref{experiment} , the upper left and lower right parts of the ground truth in Fig. \ref{groundtruth} are blank, but the generated images supplement missing parts.
Image super-resolution refers to the restoration of a high-resolution image from a low-resolution image or image sequence. Both of these tasks need to infer more information from their own information. A lot of works from neural network try to solve the problem. Dong \etal~\cite{Dong_2016} first propose a deep learning method for single image super-resolution. Ledig \etal~\cite{Ledig_2017} combine the perceptual loss and adversarial loss to train the network to obtain more realistic high-resolution images.

Furthermore, it is important to learn interpretable decomposition representations from network to achieve image editing.
Higgins \etal~\cite{Higgins2017betaVAELB} introduce $\beta$-VAE, a new framework for automated discovery of interpretable factorised latent representations from raw image data in a completely unsupervised manner.
Ricky \etal~\cite{chen2018isolating} inspire $\beta$-TCVAE (Total Correlation Change Auto-Encoder) to learn the untangled representation, thereby achieving controllable facial editing to a certain extent. 
Pan \etal~\cite{pan2020exploiting} achieve general image restoration and manipulation by learning a good long-term prior of image. As shown in Fig. \ref{image_editing}, Liu \etal~successfully realized user interactive visual content filling into the image editing program
These studies are closely related to the research of general image transformation.

\subsection{Image and Video Forgery about Face}
Portraits, especially facial shots, have a significant proportion of the images. Many applications about human faces, such as face recognition, facial expression recognition, etc., are widely used in society. The image transformation discussed earlier is mostly for general images. For facical images, there are also massive studies to improve the aforementioned general methods.
Derived from the general image translation, I. {Korshunova \cite{8237659} proposed the face-swap using similar network and loss, which realized the replacement of different faces in the same image in Fig. \ref{face-swap}.
Face-swap can also provide social communication and entertainment.
TikTok \cite{weimann2020research} realizes the replacement of faces in one video on mobile phones. which has attracted a huge audience of $1.5$ billion active users. 

These technologies make it difficult to identity fake images or videos.
Some malicious images and videos have been forged. For example, the images of celebrities, politicians and others have been maliciously modified to damage their reputations. The detection of tampered images is currently a hot topic, and we will introduce relevant research in Section \ref{detection} as the future challenge.

\begin{figure}
\begin{center}
\includegraphics[scale=0.63]{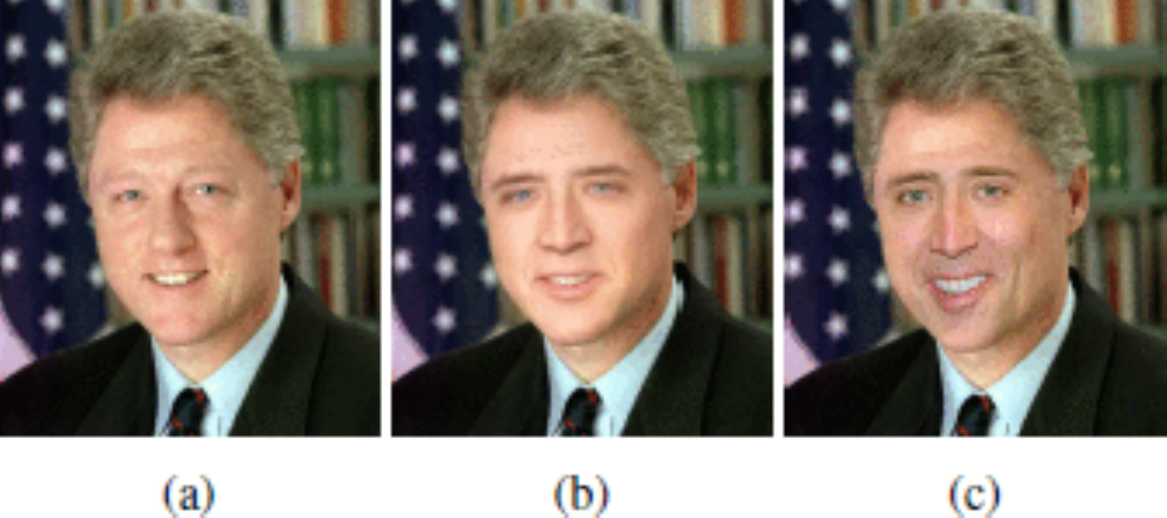}
\end{center}
\caption{(a) The input image. (b) the result of face swapping with nicolas cage using our method. (c) the result of a manual face swap (source: http://niccageaseveryone.blogspot.com). The figure comes from \cite{8237659}}
\label{face-swap}
\end{figure}

\subsection{Social Communication and Entertainment}
style transfer can create impressive image, such as learning the painting styles of famous painters like Van Gogh.
As is shown in Fig. \ref{style_image}, some excellent pictures are attractive and widely spread on social media. 
More and more people would like to implement custom style transfer for their images.
The web Ostagram \footnote{ \url{https://www.ostagram.me}}  makes it easy for anyone to accomplish style transfer by uploading two pictures. The created high-quality pictures can be publicly used for social communication like in Fig. \ref{ostagram_image}.
Prisma \footnote{ \url{https://www.prisma.io}} is a mobile application that provides similar services. TikTok attracts a huge audience of $1.5$ billion active users, mostly children and teenager as the fastest-growing application \cite{weimann2020research}.
Although most social attention is focused on leading platforms such as Twitter and Facebook, TikTok is increasingly being used by young people under the age of 13.
Use technologies to attract user participation, these applications are playing roles about social communication and entertainment. 

\begin{figure}
\begin{center}
\includegraphics[scale=0.5]{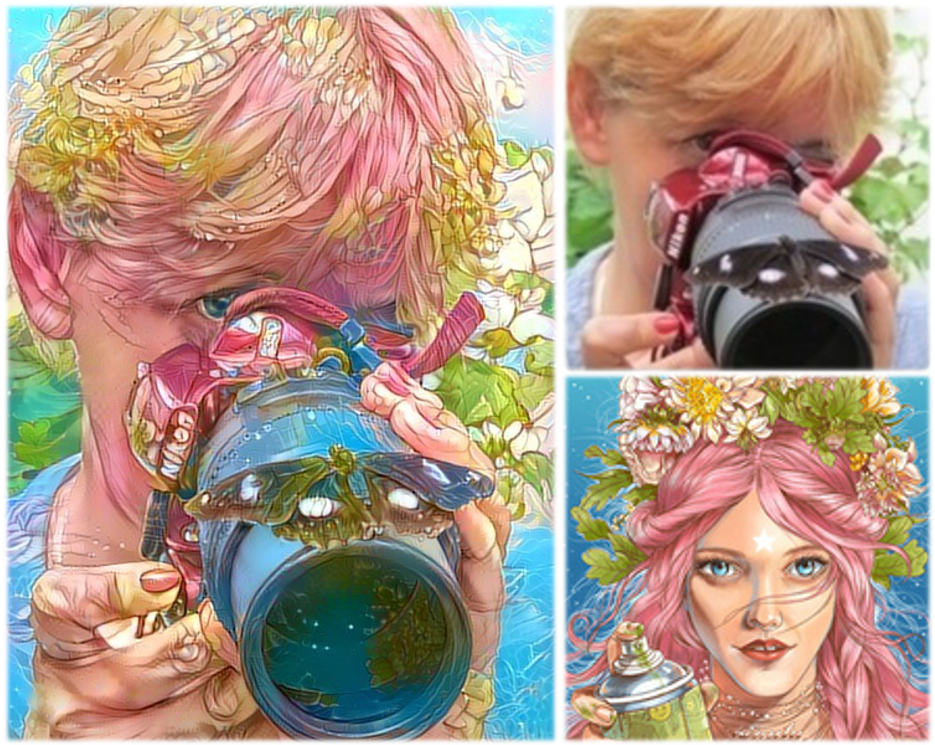}
\end{center}
\caption{The artist image combines the content image and style image from Ostagram. }
\label{ostagram_image}
\end{figure}

\section{Future Challenge}
\label{furture}

Currently, although many creative and artistic tasks about image translations have been solved by neural networks,  they extend and flourish the traditional research of image transformation. 
However, new research has brought new challenges.
First, due to the lack of data annotation, it is difficult to evaluate the results of the image transformation and style transfer. 
Second, when the technologies of image transformation end-to-end makes it easier to forge images, it is crucial to detect or classifier synthetic images. 
Third, we believe that our proposed IGC learning framework is a broad learning framework beyond the topic of image transformation.
In this section, we discuss two issues of image transformation about evaluation and detection, and broaden our proposed IGC learning framework for future research.

\subsection{Evaluation of Image Transformation}
\label{evaluation_problem}
The evaluation of the quality of synthesized images is an open and challenging problem \cite{Ruiz_2020},  because there is no correctly ground truth in some image transformation tasks.
Specifically, most evaluation methods of style transfer algorithms are qualitative. For example, the usual evaluation process is to recruit some users from Amazon Mechanical Turk. The synthesized and real images are judged by humans. The evaluation process by humans is subjective and costly. The results vary with different observers. Therefore, the auto evaluation by the program is the current direction of research.


For the general image transformation methods, we evaluate them from a variety of perspectives. According to the student model, we can compare its size and computational cost. The efficiency of inference is the most concerned in the deployment phase, especially on resource-constrained devices, e.g., mobile devices. Multiple-domain transformation methods improve the efficiency of image-to-image translation among objects. As for evaluating the quality of generated image, we can utilize the ground truth on aligned dataset. The $\mathcal{L}_{pixel}$ and $\mathcal{L}_1 $ are used to measure the difference between the generated image and the label. However, it's not good metrics to evaluate the synthetic images, which we experiment in Section \ref{experment1}. Furthermore, the metrics do not work on unaligned dataset. More and more researchers adopt the Intersection Over Union (IOU) \cite{Rezatofighi_2018_CVPR} as an evaluation metric.
IOU is to calculate relatively the intersection pixels about a certain object between the ground truth and the output of image transformation, as defined.
\begin{equation}
IoU = \frac{\text{Area of Overlap}}{\text{Area of Union}}
\end{equation}
When the general image transformation task on cooperative learning is related to object transformation, the classifier network and object detection network in the CV are used to evaluate the quality of images. 
Table \ref{resultcyc} shows the classification accuracy and IOU about CycleGAN. The ablation experiment verifies that cooperative learning improves the student model of image-to-image translation.

\begin{table}
\begin{center}
\caption{The ablation experiments of CycleGAN. The results evaluated on Cityscapes from label to photo.}
\begin{tabular}{cccc}
\hline
Loss & $\mathcal{L}_{pixel} $ & Per-class acc & Class IOU \\
\hline
Cycle alone & 0.22 & 0.07 & 0.02 \\
GAN alone & 0.51 & 0.11 & 0.12 \\
CycleGAN & 0.52 & .17 & 0.11 \\
\hline
\label{resultcyc}
\end{tabular}
\end{center}
\end{table}

For style transfer methods with guided learning, it is more difficult to quantitatively evaluate the quality of images. 
In the work of Jing \etal, they experimented style transfer methods with $10$ styles images and content images. Then, the visualization of the experimental results demonstrates that each method accomplishes the image fusion task and has its own characteristics. Although the style itself originates from subjective feelings, researchers also evaluate those methods from other aspects, such as the speed and diversity of transformation. In Section \ref{styletransfer}, we introduce some methods which deal with various content image and style images in one student model. These methods reduce the time of inference and improve the diversity of styles.  On resource-constrained mobile platforms, the consumption of resource in the model is also used as an evaluation metric. 
DIN achieves arbitrary style transfer via MobileNet \cite{howard2017mobilenets}, and reduces the computational cost by more than $20$ times compared with the existence methods.

In summary, although we can evaluate the model from the consumption of model, inference time, and diversity, we are still hard to quantitatively assess the quality of transformation images. More contributory work is worth looking forward to in the future.

\subsection{Detection of  Image Transformation}
\label{detection}
image transformation makes it easy to fake images and even videos. When it comes to the security field, it is important to detect whether the image has been fabricated.  
The Deepfake Detection Challenge (DFDC) Dataset \cite{dolhansky2019deepfake}, which contains more than $100000$ videos, is created to promote the detection of image transformation. 
Wang \etal~\cite{wang2019detecting} present a neural network method to detect the manipulation by Photoshop and apply it to real artist-created image manipulations. 
Rossler \etal~\cite{9010912} propose a method to examine the realism of state-of-the-art image manipulations, especially on the facial images. 

However, the existing detection methods rely on specific datasets and generation algorithms, which leads to the weakness of their generalization. 
Therefore, to enhance the capability of models for detection of image transformation, the models should be exposed to more image transformation methods, which helps models generalize on more transformation images. 
The traditional detection methods can learn from the guidance learning, i.e., the detection model is the student model, and the image transformation model is the teacher model. Nguyen \etal~proposed multi-task learning to detect and segment manipulated facial images and videos, which is corresponding to the individual learning of the IGC framework. The model consisted of multiple student models have stronger the generalization of detection.
Some researches on the adversarial examples provided the development of active defence algorithms \cite{Khan2021AdversariallyRD}. This kind of detection methods are designed for unknown type of transformed images. Employing adversarial examples is a typical cooperative learning method. The detection of image transformation is also a hot topic. The manipulation and detection are a zero-sum game, and the development of forgery technologies continuously launch a new round of challenge to detection field. 

\subsection{Extension of the IGC Learning Framework}
From the previous models and experiments, many models can be unified on the IGC learning model framework. Although each learning method is relatively independent, a mixed model of multiple learning methods can often be used in specific problems. We believe that the exploration of effectively learning ways among multiple models is a challenging research direction.

The paper \cite{sabour2017dynamic} proposes a new capsule network. A capsule is a group of neurons whose activity vector represents the instantiation parameters of a specific type of entity (such as an object or a part of object). They divide the different parts of the object whose each part is represented by a capsule. The capsule can been seen as a student network from our IGC learning framework. A series of capsules have their own independent capacities, and they can be connected to learn parameters by appropriate methods, e.g.,  dynamic routing mechanism in \cite{sabour2017dynamic}.  Those methods classified as cooperative learning enrich the research of the IGC framework.

Reinforcement learning \cite{8103164} differs from supervised and unsupervised learning but is contained in the IGC learning framework. In the absence of a training dataset, reinforcement learning is bound to learn from its experience, which is the study of decisions to make over time with consequences. The agent and environment in reinforcement learning. The learning methods in reinforcement learning can be regarded as guided learnings in the IGC framework.

In summary, we believe the IGC learning framework based on annotated information is a unified framework to research neural network Systematically. Massive neural network research can be incorporated into it. More works deserve to be reconsidered and analysed with the IGC framework in the future.

\section{Conclusion}
\label{conclusion}
In this article,  we reviewed the image transformation from the perspective of the IGC learning framework that we summarize and categorize the annotation to train the neural network. Based on that, we analysed some image-to-image translation methods on an aligned dataset with independent learning, the general image translation on an unaligned dataset with cooperative learning, and the image style transfer with guided learning.
In addition, we summarized the development of image translation and style transfer following one model to accomplish a single image or multiple image transformations. To our best knowledge, we are the first to propose a unified perspective called IGC learning which is adopted to analyse image transformation and style transfer about neural network image transformation. We experimented with related methods based on the learning methods of IGC on three types of datasets and analysed the advantages and disadvantages of their respective learning.
Furthermore, we discussed some popular applications and related topics. 
At the end, we proposed possible directions to be research in future works.

\bibliographystyle{IEEEtran}

\bibliography{bare_jrnl}

%








\end{document}